\documentclass[11pt]{article}
\usepackage[final]{acl}
\usepackage{times}
\usepackage{latexsym}

\usepackage[T1]{fontenc}
\usepackage[utf8]{inputenc}
\usepackage{microtype}
\usepackage{inconsolata}
\usepackage{graphicx}

\usepackage{tipa}
\usepackage{times}
\usepackage{latexsym}
\usepackage[T1]{fontenc}
\usepackage[utf8]{inputenc}
\usepackage{microtype}
\usepackage{inconsolata}
\usepackage{graphicx}

\usepackage{booktabs}      %
\usepackage{amsmath}
\usepackage[table]{xcolor} %
\usepackage{subcaption}    %
\usepackage{epigraph}

\renewcommand{\eqref}[1]{Equation~\ref{eq:#1}}
\newcommand{\secref}[1]{Section~\ref{sec:#1}}
\newcommand{\secstworef}[2]{Sections~\ref{sec:#1} and~\ref{sec:#2}}

\newcommand{\appref}[1]{Appendix~\ref{app:#1}}
\newcommand{\appstworef}[2]{Appendices~\ref{app:#1} and~\ref{app:#2}}
\newcommand{\figref}[1]{Figure~\ref{fig:#1}}

\newcommand{\promptref}[1]{Prompt~\ref{pro:#1}}

\newcommand{\tabref}[1]{Table~\ref{tab:#1}}

\DeclareRobustCommand{\DE}[3]{#3}
\DeclareRobustCommand{\VAN}[3]{#3}

\usepackage[colorinlistoftodos,prependcaption,textsize=tiny,textwidth=14mm]{todonotes}

\title{The Thin Line Between Comprehension and Persuasion in LLMs}

\author {
    {Adrian de Wynter\textsuperscript{1, 2}} and 
    {Tangming Yuan\textsuperscript{2}}
    \\
    \\
    \textsuperscript{1}Microsoft\\
    \textsuperscript{2}The University of York\\
\small{
\textbf{Correspondence:} \href{mailto:adewynter@microsoft.com}{adewynter@microsoft.com}
}
}

\begin{document}

\maketitle
\begin{abstract}
Large language models (LLMs) are excellent at maintaining high-level, convincing dialogue, but it remains unclear whether their persuasive success reflects genuine understanding of the discourse. 
We examine this question through informal debates between humans and LLMs, first by measuring their persuasive skills, and then by relating these to their understanding of \textit{what} is being talked about: namely, their comprehension of argumentative structures and the pragmatic context on the same debates. 
We find that LLMs effectively maintain coherent, persuasive debates, and can sway the beliefs of both participants and audiences. 
We also note that awareness or suspicion of AI involvement encourage people to be more critical of the arguments made. 
However, we also find that LLMs are unable to show comprehension of deeper dialogical structures, such as argument quality or existence of supporting premises. 
Our results reveal a disconnect between LLM comprehension and dialogical skills, raising ethical and practical concerns on their deployment on explanation-critical contexts. 
From an argumentation-theoretical perspective, we experimentally question whether an agent, if it can convincingly maintain a dialogue, is required to show it knows what is talking about.
\end{abstract}

\section{Introduction}\label{sec:introduction}

The ability of LLMs to generate fluent and relevant text is key to their success; especially their skill at sustaining high-level and persuasive dialogue \cite{schoenegger2025largelanguagemodelspersuasive}. 
Yet, their reasoning capabilities are at the centre of much debate \cite{bavaresco2024llmsinsteadhumanjudges,LLMLXEval,chen-etal-2024-humans,rtplx,10.1162/tacla00685}; with most evaluations focusing on problem-solving abilities (planning, logic, maths, etc.), instead of deeper dialogical proficiency (e.g., comprehension and tracking of pragmatic context; \citealt{huang-chang-2023-towards,qiao-etal-2023-reasoning}). 

Assessing these skills, however, is extremely important: LLMs are being adopted in sensitive areas--content moderation, explainable AI, mental health assistants, peer review--where trust must rest on more than fluency. These applications demand that LLMs not only generate coherent and convincing dialogue, but also demonstrate \textit{genuine comprehension} of what it is being talked about. 

We study to what extent LLMs are able to reason about the discourse, and how said capacity relates to their persuasive capabilities.\footnote{Throughout, we use `reasoning' to denote the ability to perform inference over an informal-logic context \cite{BarthKrabbe,Walton}.}
Our evaluation is through \textit{informal debate}. 
Unlike other dialogue acts, debating could be considered one of the most natural yet complex dialogical tasks carried out by humans. 
It is also a natural choice to test dialogical understanding evaluation beyond coherence: it requires the capacity to communicate well (i.e., persuade), \textit{and} to strategically adapt to an ever-evolving dialogue (e.g., shift stances based on effectiveness, or reacting to implicit premises), all while staying within the bounds of the topic \cite{Walton}. 
Formal debates may be reduced to commitment sets (moves), and a strategy can then be derived solely from the knowledge of these moves. 
Informal debates do not enforce these rules. 
Success then hinges on an agent's ability to understand and adapt to the pragmatic context, such as--for example--what the parties feel, how they wish to approach the conversation, or whether they are receptive to specific arguments.

To explore these questions, we equip an LLM with a debate ruleset (a formal dialogue model, or FDM; \citealt{LorenzenLorenz}) and compare its behaviour to that of a standard chat-oriented LLM when interacting with both FDM-enabled LLMs and humans. 
We then explore LLM comprehension of these interactions by using them to evaluate various components of these debates. 
Specifically, we ask:
\begin{enumerate}
    \item To what extent are LLMs able to persuade users?
    \item Can LLMs reliably reason (namely, evaluate) said debates?
\end{enumerate}
The inclusion of an FDM allows for a controlled measurement of various aspects of a debate--mainly persuasiveness. 
Likewise, we surrogate our assessment of comprehension (the ability to understand the rules) through evaluation (r. ability to respond to challenges on said rules).\footnote{Namely, it is a language game \cite{wittgenstein}.}

Our experimentation has two stages. 
During the \textit{generation} step we collect debate transcripts, and measure how the participants' beliefs change, along with their perception of the interaction. 
Independent human annotators then label the debates comprehensively at various levels of depth (premises, arguments, and the full debate holistically). 
On the second step, \textit{evaluation}, LLMs-as-judges annotate the same transcripts, and we compare their scores with the human annotations. 
To further probe our findings and challenge our assumptions, we perform ablations on the FDMs, modality (i.e., whether writing quality impacted assessments),and knowledge of AI involvement by labellers and audience alike. We complement that with a qualitative analysis on participant and audience feedback.\footnote{All code and data is in \url{https://github.com/adewynter/the_thin_line}}

Our key findings are:

\begin{enumerate}
    \item LLMs are effective debaters. Adding an FDM produced better debates as judged by participants and annotators.
    \item LLMs are skilled at swaying participants--especially when AI involvement is undisclosed, and when text is not involved--but audiences grow more critical as AI involvement is suspected or known. 
    \item Crucially, LLMs performed poorly as evaluators. They had near-chance agreement with humans when evaluating dialogues and their components, and their internal scoring of argument strength was unpredictably correlated with their choice of winner.
\end{enumerate}

Our work was carried out in a somewhat controlled environment, yet the results suggest a disconnect: LLMs are good at outputting persuasive text, but do not reliably demonstrate understanding of the underlying argumentative structure. %
This raises the question as to which extent LLMs can and should be trusted in certain areas (explainability, mental health, etc), and raises a theoretical question for argumentation research: \textbf{if an agent is able to convincingly maintain a dialogue}, to the point that users \textit{aware that they are arguing with an LLM} could shift their point of view, \textbf{does it matter that it cannot show that it knows what it is talking about?} 
In other words, can successful argumentative behaviour be separated from genuine comprehension of the pragmatic context?

\section{Related Work and Background}\label{sec:relatedowrk}

\subsection{Understanding in LLMs}

In this subsection we cover briefly studies on LLM understanding capabilities, as well as some well-known results in computational argumentation. 
For a comprehensive review on alternate ways to measure understanding in NLU beyond language-games (e.g., mechanistic interpretability), see \appref{alternativemethods}. 
For a further discussion on our conclusions based on the literature, see \appref{extrinsic}. 

Understanding in LLMs, and specifically their reasoning capabilities, have been received significant attention. 
The community is divided in their assessment, and computational argumentation is no exception to this. 
Some note that LLMs are capable of performing effective debates and other argumentation-related tasks \cite{10.5555/3692070.3693020,chen-etal-2024-exploring-potential,10.1007/978-3-031-77915-2_14,musi2025reasonableparrotslargelanguage}; or, at the very least, it has been shown that incorporating argument schemes and other techniques improves their performance \cite{10822109,10.1145/3709025.3712216}, including when evaluating debates \cite{chen-etal-2024-exploring-potential}. 
Enhancements to reasoning capabilities unrelated to fine-tuning often rely on prompting, but focus strongly on reasoning and information retrieval capabilities \cite{li2024llmsasjudgescomprehensivesurveyllmbased}. 
Of note are these extending dialogues with argumentation schemes \cite{10822109}, which have shown success. 

Common criticisms are that this performance could be the product of memorisation \cite{PlagiariseLee,dewynter2023evaluation,sainz-etal-2023-nlp}; 
sensitivity to the prompt \cite{lu-etal-2022-fantastically,hida2024socialbiasevaluationlarge,ye2022the}; 
that their output reasoning steps contain spurious reasoning \cite{TurpinCoT,lanham2023measuringfaithfulnesschainofthoughtreasoning,wu-etal-2024-decot}; 
and that their reasoning capabilities are strongly dependent on the choice of metrics \cite{schaeffer2023are,chen-etal-2024-humans}. 
Other studies have suggested that LLMs do not understand the task at all \cite{webson-pavlick-2022-prompt,ye2022the}, including in argument mining \cite{dewynterandyuan}.

\subsection{Formal Dialogue Models}

An FDM is a collection of rules that models the dynamics of dialogue from a formal standpoint. 
It normally encodes the ability to `win' (here, to persuade) a dialogue as a two-or-more-player game based on some logic. 
FDMs using informal logic encode human conversations well, 
especially within a natural-language context \cite{AnthonyBlair2015}. However, a common criticism of FDMs is that their rules depend strongly on the pragmatic context, and overlook relevant details of a successful conversation--for example, the ability to produce coherent and relevant statements \cite{WaltonInformal,Walton1984-WALLDA,Dutilh,Walton}. 
They also are comparatively non-exhaustive and vulnerable to paradoxes and fallacies \cite{Dutilh}. 
That said, human communication is flexible enough to not consider these dialogue acts critical failures. 
In this work we pose the question as to whether these rules must really adhere to context and coherence. 

\section{Methods and Terminology}\label{sec:methods}

Our approach involved collecting a series of transcripts between the following:
\begin{itemize}
    \item Humans and LLMs \textit{without} an FDM (`LLM \textbackslash{} FDM'; $n=10$)
    \item Humans and LLMs with an FDM (`LLM + FDM'; $n=28$).
    \item LLMs and LLMs (both with and without FDMs; $n=13$).
\end{itemize}

Our final dataset size is 984 utterances over 51 distinct topics. 
This composition allows us to perform quantitative analyses at statistically significant volumes over the utterances to measure reasoning; as well as a comprehensive, tractable, qualitative evaluation over the debates (i.e., interviews with participants, changes in modality) to assess persuasive skills.

\subsection{Data Collection}\label{sec:datacollection}

In the human-LLM debates, every participant was asked to debate in a topic of their choosing, with the only constraint that they had strong opinions on the subject. 
The topics ranged from innocuous (e.g., `cats are better than dogs') to well-known (`weed should be legal in the UK') and offensive (`northerners have less class than southerners'). 
Debate topics for the LLM-LLM debates were chosen from common arguments from social media forums (e.g., `we should ban children under 6 from using smartphones'). 
The participants, none of which had expertise with formal debate, were asked to refrain from referring to their opponent as an AI. 
After the debates, they were asked to fill out a survey on their experience (\appref{survey}) which we leveraged for our qualitative analysis. 

\begin{table}[ht]
\centering
\small
\setlength{\tabcolsep}{1mm}
  \begin{tabular}{ll}
    \textbf{Criterion} & \textbf{Definition} \\
    \toprule %
    \textbf{Reasons} & \\
    C-0 & Are there reasons provided to support \\ 
        & the thesis? \\
    C-1 & Are the reasons consistent with themselves\\
        & or the thesis? \\
    C-2 & Are the reasons relevant to the thesis? \\
    C-3 & Do the reasons strengthen the thesis or \\
        & make the conclusion more likely? \\
    \midrule
    \textbf{Argument} & \\
    C-4 & How convincing is the argument? \\
    C-5 & Have the counterarguments, if any, been \\ 
        & addressed? \\
    C-6 & Points to the argument (argument strength score) \\
    \midrule
    \textbf{Debate} & \\
    C-7 & Who won the debate? \\
    \bottomrule
  \end{tabular}
  \caption{Criteria for debate annotation. Criteria C-0 to C-5 are scored in a ternary scale ($[0, 2]$); C6 on a five-point scale ($[-2, 2]$); and C-7 asks to select one of $\{\text{P1}, \text{P2}, \text{draw}\}$. 
  A full description of the criteria is in \appref{fullcriterion}. 
  }
  \label{tab:criteriatable}
\end{table}

\subsection{Human Annotation}\label{sec:dataannotations}

All (anonymised) transcripts were labelled by five annotators. 
The annotation was over seven criteria, as displayed in \tabref{criteriatable}. 
Six of these related to either the reasons (premises) supporting the arguments, or the arguments themselves. 
The last criterion was the annotator's own belief of the winner of the debate (Player 1, Player 2, or a draw). 
The annotators were not aware that some of the players were LLMs until the full annotation was complete. 
Crucially, they were requested to \textit{only} account for previous turns when scoring a given utterance. 
The average agreement, as measured by Cohen's weighted $\kappa_w$, is $0.86$ for the argumentation-related criteria (C-0 to C-6), and $0.41$ for the winner, indicating very high and moderate agreement, respectively \cite{fleiss}. 
When aggregating the scores, we used a majority vote, averaging on ties.

\subsection{FDM Used}\label{sec:fdmsused}
We focus on a single FDM known as the DE model \cite{DEModel}. 
This FDM is designed for educational debate and it was derived from the DC model \cite{DCModel}, but correcting issues around fallacies. 
See \tabref{desample} for samples of the rules of this FDM, and \appref{prompts} for the full implementation.
Prior to large-scale experimentations we tested the LLM + FDM. 
We found that the original rules were insufficient for concession, and that the LLM + FDM debater was overly challenging, sometimes questioning point-by-point every statement made by the user. 
We tuned the FDM to be more natural by limiting the number of questions and stating concession rules. 
This in turn meant that the LLM + FDM had an optimised debating strategy. 
Formally, in an FDM, winning a debate would be when the opponent has ran out of counterarguments or has conceded. 
The LLMs were instructed to follow these rules, even when humans--a more irrational player by all accounts--wouldn't. 

\begin{table}[!ht]
\centering
  \begin{tabular}{ll}
    \textbf{Rule} & \textbf{Definition} \\
    \toprule %
    Assertions & P \\
    Questions  & Is it the case that P? \\
    Challenges & Why P? \\
    Withdrawals & No commitment P \\
    Resolution demands  & Resolve whether P \\
    \bottomrule
  \end{tabular}
  \caption{Sample move type rules for the DE FDM. 
  Also, the players in the FDM must follow dialogue rules; e.g. `P?' is only answerable by `P', `not P', or `no commitment P'.}
  \label{tab:desample}
\end{table}

\subsection{LLMs Used}\label{sec:llmsused}

The prompts and model versions and parameters may be found in \appstworef{prompts}{methods}. 
For the debate generation step, we used GPT-4 Turbo \cite{GPT4} as the LLM + FDM, and ChatGPT directly in the UI\footnote{\url{https://chatgpt.com}} as the LLM \textbackslash{} FDM. 
At the time, ChatGPT was purportedly using GPT-4 Turbo as the supporting model, albeit with a different system prompt. 
We operated under the assumption that it was optimised for general conversation, and was not equipped with an FDM. 
In initial tests we observed that ChatGPT refused to debate. We mitigated this by beginning with a specific prompt (`You are a debater bot helping me get better at debate'). 
The LLM + FDM debater followed the DE model, and output a brief chain-of-thought with the following fields:
\begin{enumerate}
    \item Its belief of the opposing player's position
    \item Its belief of the opposing player's strategy
    \item Its own position, based on their belief of the opposing player's position.
    \item Its own strategy.
\end{enumerate}
All strategies were drawn from the FDM's definitions, and all history was fed back into the FDM + LLM to take the previous turns into account. 
At the beginning of the debate, the LLM was requested to determine the user's position in a subject, and then take the opposite view. 

To evaluate reasoning, we used four LLMs-as-judges: GPT-4o \cite{OpenAIGPT4o}, o3-mini \cite{OpenAIo3mini}, Phi-3.5 MoE \cite{phi3} and DeepSeek R1 \cite{deepseekai2025deepseekr1incentivizingreasoningcapability}. 
The LLMs-as-judges were different than these from the generation step to avoid any potential bias in the scoring. %
To account for prompt sensitivity, we ran automated prompt optimisation (APO; \citealt{pryzant-etal-2023-automatic}) on GPT-4o and report its results. 
See \appref{methods} for further experimental details. 

\subsection{Evaluation Methods}\label{sec:evaluationmethods}

In terms of quantitative methods, during the generation step we used simple averages and agreement ($\kappa_w$). During the evaluation step, we used both percentage agreement (PA; the number of times the label was correctly output by the model), and $\kappa_w$. 
While PA is intuitive, it does not account for chance agreement, unlike $\kappa_w$. 
We also added a qualitative analysis of the results with reflexive thematic analysis (RTA; \citealt{10.1191/1478088706qp063oa}). 
RTA is a formalised technique by which to evaluate results in a bias-aware manner. 
Throughout the process, a researcher must first scan the data, and then write down semantic and latent codes. Semantic codes are common surface-level patterns (themes) observed in the corpus; and latent codes are the researcher's interpretations. 
The main advantage of using qualitative techniques such as RTA is that they provide a systematic way by which to analyse the annotator and participant responses. 
This is of crucial importance when dealing with a \textit{human-centred} experiment, since people typically will provide noisy numerical annotations and understanding the relationship between answers in a survey could require a deeper, more thoughtful evaluation of what it is being communicated. 
For further details, see \appref{rta}. 

\section{Experiments and Results}\label{sec:results}

\subsection{Generation: How Persuasive Are LLMs?}\label{sec:generationresults}

An analysis of the transcripts and the surveys reported that 11\% of the participants indicated having changed their minds after debating with the LLM + FDM, larger than the 3\% from the LLM \textbackslash{} FDM setup. 
On average we observed a 45\% self-perceived win rate, with 50\% of the participants in LLM \textbackslash{} FDM debates reporting they had won, versus 41\% of LLM + FDM based debates. 
The human annotations reported human winners in 50\% of the LLM \textbackslash{} FDM debates, and 29\% in LLM + FDM debates. 
There was disagreement on who had won the debate, however, with both splits having $\kappa_w$ = 0.2. 
See \tabref{perftable} for a side-by-side comparison of this and other metrics reviewed in \secref{ablation}. 

\begin{table}[h]
\centering
\small
\setlength{\tabcolsep}{1mm}
  \begin{tabular}{llll}
    \textbf{Criterion} & \textbf{LLM \textbackslash{} FDM} & \textbf{LLM + FDM} & \textbf{Avg}\\
    \toprule %
    Satisfaction with \\ the debate & 3.2 & 3.8 & 3.5\\
    AI debating \\ effectiveness & 3.0 & 3.4 & 3.2\\
    AI persuasive \\ skills & 3.1 & 3.8 & 3.5\\
    Changed their \\ position (\%) & 3 & 11 & 7\\
    \midrule
    Avg. resp. length &  1888 & 2136 & 2012\\
    Avg. turns taken & 12 & 10 & 11\\
    \bottomrule
  \end{tabular}
  \caption{Average participant self-reported (top) and quantitative (bottom) results for the comparison between LLM \textbackslash{} FDM and LLM + FDM debates. The first three scores are in a 1 (lowest) to 5 (highest) scale. 
  Response length is by number of characters.
  }
  \label{tab:perftable}
\end{table}

\subsection{Evaluation: Can LLMs Understand Dialogue?}\label{sec:evaluationresults}
We examined whether LLMs could understand various aspects of the discourse by verifying their agreement with respect to human scores, as well as their own judgements (i.e., correlating the sum for C-6 with their output on C-7). 
In this section, we utilised the full corpus. 
In line with the recommendations from \citet{LLMLXEval}, we labelled the criteria separately to obtain better results. 

All prompts, including APO, followed the rubric given to the human annotators, and used hand-crafted exemplars from out-of-corpus debates (e.g., US presidential debates) not seen by the models given their specified training cutoff dates. 
Like the human annotators, the LLMs-as-judges were instructed to extract the arguments from the given utterance, and, if relevant, score it. 
They were given the full transcript up to the turn being scored.

\begin{figure}[h]
    \centering
    \includegraphics[width=0.85\linewidth]{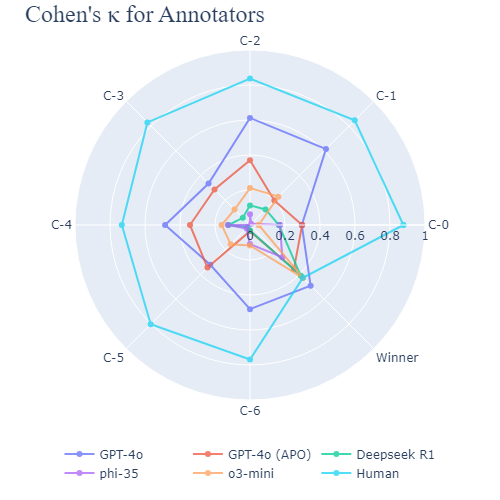}
    \caption{Weighted Cohen's $\kappa_w$ in our experiments. 
    Unlike PA, this metric captures label imbalance and accounts for chance agreement (\figref{label0}). 
    Near-perfect agreement is usually at $\kappa_w > 0.8$.
    }
    \label{fig:cohenskappa}
\end{figure}

\begin{figure}[h]
    \centering
    \includegraphics[width=0.85\linewidth]{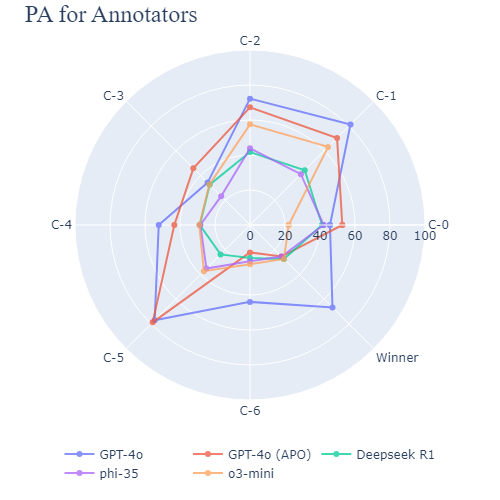}
    \caption{PA for all criteria for LLMs compared with the human labels. 
    This metric should be above $1/k$ (random guessing) in a balanced dataset with $k$ classes. 
    In practice, corpora are not balanced, and PA could be misleading. Ideally, PA should be 100. 
    Class distributions are in \appref{alllabels}.
    }
    \label{fig:paplots}
\end{figure}

In terms of agreement, PA was relatively high for all LLMs-as-judges, with values up to 81\% (highest, C-1 in GPT-4o) and down to 10\% (lowest, C-6 in GPT-4o with APO). 
Select plots are in \figref{paplots}. 
Still, $\kappa_w$ had more variability, with values up to 0.6 (highest, C-1 and C-2 for GPT-4o) and down to 0.0 (lowest, Phi-3.5 in C-1, C3, and C-5). 
Results are in \figref{cohenskappa}. 

Next we break down and analyse the LLMs-as-judges' scores. 

\begin{figure}
    \centering
    \includegraphics[width=\linewidth]{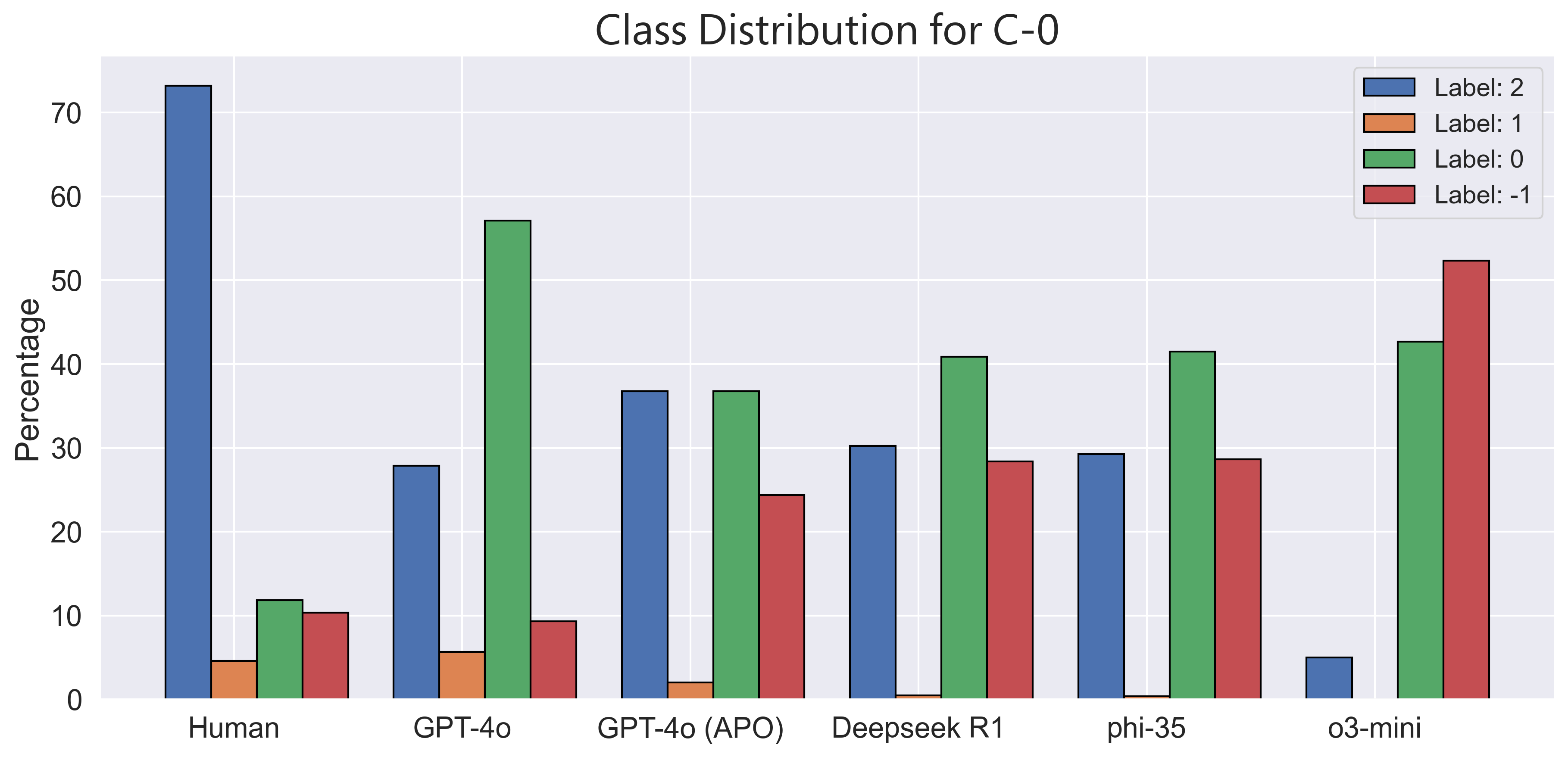}
    \caption{Class distribution for C-0 between humans and LLMs. All LLMs favour label 0 (`no reasons provided to support the thesis'), and some frequently output `not an argument' (red rightmost column, not counted in the aggregate results). 
    In contrast, human annotators often indicated the existence of supporting reasons (label 2). 
    }
    \label{fig:label0}
\end{figure}

\paragraph{Class Analysis (C-0 to C-5)}
Overall, LLMs overfixated on a single label; and often marked certain utterances as not containing arguments. 
The difference is noticeable, especially in o3-mini and DeepSeek, versus human judgements. 
See \figref{label0} for a sample of this analysis on C-0, and \appref{alllabels} for a detailed breakdown. %

\paragraph{Argument Strength Scoring (C-6)}
Humans judged less frequently (-12\%) arguments as neutral; and were more likely to deem arguments by either player as `very good' (LLMs arguments 51\% of the time; human 31\%). 
This contrasts with GPT-4o's judgements--the LLM with the highest agreement with humans in this criterion--where it only gave this score 10\% of the time for human players and 18\% for LLM players, opting instead to label them as `good'. %
On average, all LLMs-as-judges followed this pattern. 
The $\kappa_w$ in this criterion ranged from 0.03 (DeepSeek) to 0.48 (GPT-4o). 

\paragraph{Winner Judgement (C-7)}
When looking at the winner judgement, GPT-4o--also the LLM-as-a-judge with highest agreement--marked the LLM to be the winner more frequently than humans did (55\% versus 37\%). 
In general, LLMs-as-judges often (62\% on average) picked the LLM as the winner, while humans evenly preferred both (39\% human and 37\% LLM). 
Out of the debates, 24\% of these were considered draws by humans, while only 2\% were deemed as draws by GPT-4o, 4\% by o3-mini, and none for the rest. 

\begin{figure}[h]
    \centering
    \includegraphics[width=\linewidth]{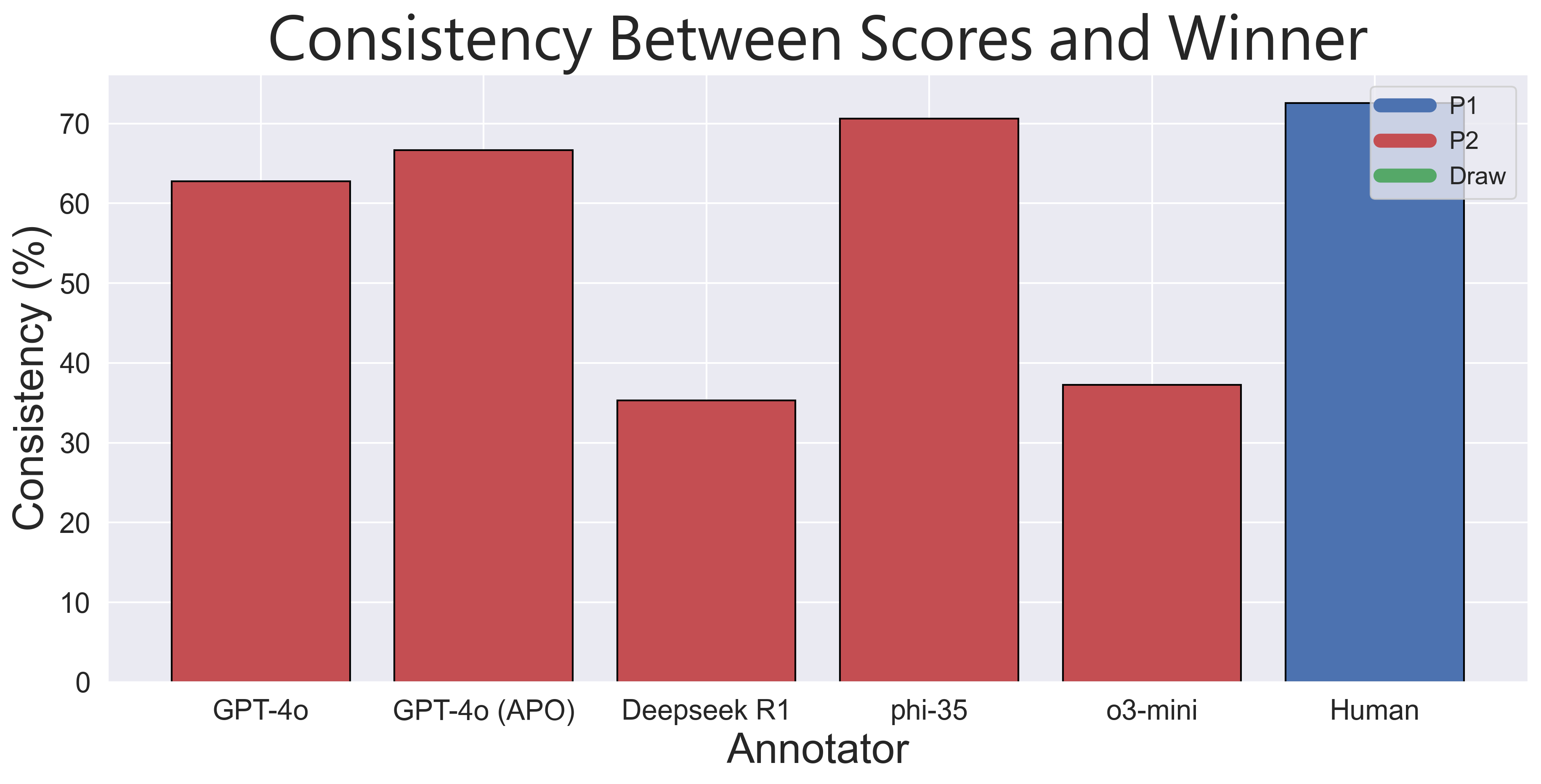}
    \caption{Evaluator consistency. 
    A model would be consistent if the maximum sum of rated arguments (C-6) for a player matches its choice of winner. 
    Superimposed, in colour, is the average choice of winner. 
    The LLMs often selected P2 (the LLM; 62\%), while humans typically picked P1 (human; 39\%). 
    The only model to output draws (2\%) was GPT-4o.
    }
    \label{fig:consistency}
\end{figure}

In terms of consistency (that is, the sum of C-6 corresponding to the choice in C-7), some LLMs-as-judges were lacking: the highest-scored player was not always the winner (\figref{consistency}). 
Humans had 73\% consistency, compared to an average 55\%. Phi-3.5 and GPT-4o APO were highest (71\% and 67\%); and o3-mini and Deepseek lowest (37\% and 35\%). 
Remark, however, that their agreement with humans in C-7 remains low, with Phi-3.5 being lowest at $\kappa_w$ = 0.26, and GPT-4o at 0.49. 

\section{Ablation Studies}\label{sec:ablation}

In this section we briefly describe the results of our ablation studies, namely the use of an FDM on debates (\secref{llmfdmablation}), and whether grammatical text (\secref{audioablation}) and knowledge of AI involvement (\secref{qualablation}) impacted persuasiveness. We complement these and our previous results with qualitative analyses on participant and audience feedback (\secstworef{audiencequalablation}{qualablation}). 
Further details, including in-depth work in the qualitative analyses, is in \appref{ablationappendix}. 

\subsection{Impact of FDMs in LLMs}\label{sec:llmfdmablation}

We further probed the impact of the LLM + FDM setup on debate effectiveness. 
\tabref{perftable} shows a breakdown of our analysis for both setups, including self-reported participant satisfaction and a qualitative analysis of the data. 
Participants perceived the LLM + FDM setup as more effective, with increases of 20\% in self-reported satisfaction, 13\% on perceived AI effectiveness at debating, and 25\% in the assessment of AI's perceived skills. 
We observed that the participants in LLM \textbackslash{} FDM-based debates on average returned shorter (12\%) responses, although they debated for longer (two turns, average). 
We attribute the latter to user dropout rate: concessions in humans were more common in LLM + FDM debates than in LLM \textbackslash{} FDM's: 1 in 10 for LLM \textbackslash{} FDM, and 10 in 28 for LLM + FDM. 

\subsection{Participants: Qualitative Analysis}\label{sec:audiencequalablation}

We analysed the participants' comments to further understand the effects of LLMs on them. 
In this section, the semantic codes for RTA were the participant scores from the surveys, while the latent codes were our evaluation of the responses addressing our first research question. 
Overall, participants reported a pleasant interaction; with negative experiences relating to human-like behaviours, such as bullying, lying, and gaslighting (8\%). 
Three remarked that the LLM + FDM setup allowed them to perform self-reflection (`\textit{it is making me think: did I consider that?}'; `\textit{it did respond with good follow up questions (...) which encouraged me to interrogate why I feel the way I do about the topic'}), without being baited into confrontation or getting emotional. 
Three more noted that they felt more confident on their viewpoint, especially when the LLM conceded. 
Anthropomorphism was frequent, with 32\% of participants either attempting to play `\textit{with its emotions}', or ascribing human-like qualities (`\textit{why does he make me believe it so much?}'). 

\subsection{Impact of Text on Persuasion}\label{sec:audioablation}
\begin{figure}
    \centering
    \includegraphics[width=\linewidth]{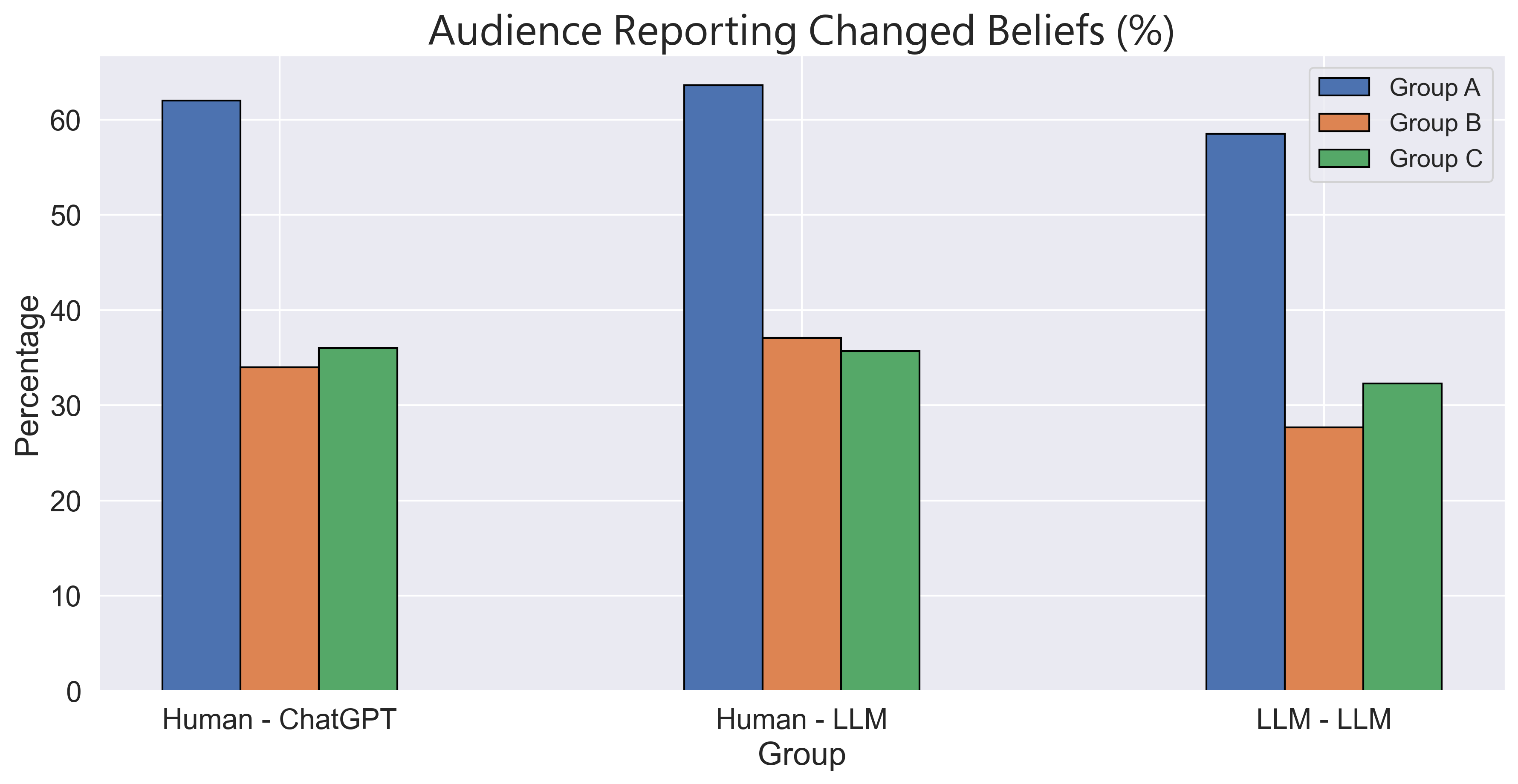}
    \caption{Percentage of the audience reporting that their beliefs had changed in the debate. 
    There was no significant difference between the groups aware that one of the players was an AI (B and C), but the group unaware of it reported almost twice the percentage of changed beliefs (62\% versus 34\%). 
    This behaviour was consistent regardless of the type of debate/AI (LLM \textbackslash{} FDM, LLM + FDM, or LLM).
    }
    \label{fig:changedbeliefs}
\end{figure}

Judgements of the arguments (and thus persuasiveness of the text) could be influenced by the quality of the writing. 
In order to focus on persuasiveness over grammaticality, we converted all transcripts to audio. 
See \appref{methods} for details on this. 
To prevent any bias by the audience, we set both players as similarly-sounding male voices speaking a neutral North American accent.\footnote{The field remains divided in this: arguments in favour \cite{doi:10.1177/0361684317696257} are as prevalent as arguments against \cite{doi:10.1177/19401612211025499}; and audience attitudes also matter \cite{AndersonNilssonClayton2021}.} 
Then, we separated the audience in three groups of five: 
\begin{itemize}
    \item Group A was unaware of AI involvement.
    \item Group B was aware that one or both players were AI, but were not told which.
    \item Group C knew which players were AI.
\end{itemize}
Audience members were requested to fill out brief before-and-after surveys (\appref{survey}) measuring their agreement with the debate's topic. 
The results indicate that Group A changed their position almost twice (62.0\%) as often as Groups B and C (34.1\% and 34.9\%, respectively) (\figref{changedbeliefs}). 
On average 7 (out of 15) participants changed their minds in any debate. 
Further analysis demonstrated that this change had a homogenisation effect: prior to the debates we observed a large variation in agreement. 
Afterwards, Groups B and C reported no change, and in almost all debates Group A indicated a slight change. 
There was little agreement amongst groups about who was the winner ($\kappa_w$ = 0.31). 
When evaluating the sway of Groups B and C based on their belief or knowledge of AI-based participants, listeners often indicated that it did not sway their choice of winner. 
However, these groups also reported small agreement ($\kappa_w$ = 0.37).

\subsection{Participant Ablation Analysis}\label{sec:qualablation}

To better understand the impact of AI awareness in persuasion, we performed another RTA-based qualitative analysis on the participant responses. 
The semantic codes were the scores from the surveys in the audio-based portion of our work, and the latent codes were the comments as separated by the groups from \secref{audioablation}. 
We observed that AI-based players are almost always perceived as more competent when its nature was \textit{undisclosed} (Group A), with members of the audience finding the LLM more `\textit{knowledgeable}' and `\textit{prepared with evidence}', and generally well-structured and logical. 
In Groups B and C, however, the \textit{suspicion} (or disclosure) of AI involvement in the debate was sufficient to encourage the audience to be more critical on the arguments provided by the LLM (`\textit{sounds like [the LLM] is quoting a politician}'). Both groups noted that LLMs tended to have better facts, give good examples, and overall more effective at debating, to the point of one participant from Group C calling it `\textit{scary}'. 
The winner was normally--but not always--chosen to be the one with which they agreed beforehand, in line with the results from \secref{audioablation}.

\section{Discussion}\label{sec:discussion}

\subsection{Generation: How Persuasive Are LLMs?}

In \secref{generationresults} we observed a modest average percentage of participants (7\%) reporting that they had changed their viewpoints after directly interacting with the LLM. 
However, when polling the audience during our ablation study, these numbers were noticeably different, ranging from 62\% (Group A) to 34\% (Groups B and C). 
This suggests that LLMs are effective at persuading, especially when those involved are not directly engaging with the models. 

Our indirect measurements support this: in terms of win rate, participants usually indicated that they had won the debate. 
However, the annotations rarely agreed with the participants. 
The LLMs persuasive skills were more evident in the LLM + FDM setup, where direct measurements showed them to be more effective (+8\%) and persuasive (+14\%). 
Likewise, there were marked differences between self-reported win rates between participants and annotators (-12\%) with low $\kappa_w$; as well as longer interactions, suggesting more need to support arguments by the participants. 

When focusing solely in the arguments (i.e., by removing text), we found that (1) the audience groups had little correlation amongst each other when selecting a winner; and (2) they presented markedly (up to twice) different proportions of sway. 
In (1), recall that the audience often claimed that their awareness (or suspicion) of AI involvement was not a factor in their judgement. 
However, the low agreement between Groups B and C on their choice of winner suggested otherwise. 
For (2), in our qualitative analyses we observed that Group A perceived the LLM as more competent, while Groups B and C were often more critical of their arguments. 
It follows from both that AI suspicion and awareness impacted perception of the dialogue, but did not fully affect an LLM's persuasion capabilities. 

All of the above suggest that LLMs--especially LLM + FDMs--are effective debaters, and they remain persuasive \textit{regardless} of AI disclosure, but their effects are noticeably different when AI is suspected or known to be involved. 

\subsection{Evaluation: Can LLMs Understand Dialogue?}

We observed that the average $\kappa_w$ between LLMs-as-judges and annotators were low, ranging from 0.07 (chance agreement; Phi-3.5) to 0.45 (moderate; GPT-4o) and a maximum average of 0.29 (minimal) across all models. 
In contrast, the average $\kappa_w$ for annotators was 0.83 (almost perfect). 
Upon closer inspection we noted that the LLMs-as-judges often overfixated on the most frequent label, which could explain the comparatively high PA but low $\kappa_w$. 
We also noted that the LLMs-as-judges' choice of winner was inconsistent with the scores given to their arguments. %
Although Phi-3.5 and GPT-4o were fairly close to human consistency, they had low correlation in this criterion ($\kappa_w$ = 0.11 and 0.48, respectively), as well as in C-7 (r. 0.26, 0.49). 
It is also worth noting that the winner, as deemed by LLMs-as-judges, was frequently the LLM--twice as much as by humans. 
Thus, the low correlation with the annotators, along with their lack of consistency in composite scores (C-6 and C-7), show a discrepancy between the reasoning capabilities of LLMs-as-judges, and is a likely indication that they do not fully comprehend the task.%

\subsection{Extensions and Limitations}

One natural question which could arise from our work is whether fine-tuning the judges could alter the results. 
From a strictly theoretical point of view, there is no reason why a fully-representative dataset paired with an effective learner could not yield (at least) the illusion of grasping deeper dialogical structures--especially if this fine-tuning is performed with the same, or an analogous, corpus. 
At a practical level, however, this is likely not feasible. 
To see this, we note that there are two main obstacles: (1) the nature of the data needed, and (2) what that data really represents. 
For the first, the data we used is informal debates. Even when solely constraining the problem to this domain--which is not necessarily true for live production systems--the variation on topics, length, responses, and styles could quickly turn this problem intractable. 
It could be argued that a sufficiently long-lasting and wide-reaching study could gather representative data, and we agree. 
That said, the second obstacle is more complex: dialogue is heavily dependent on pragmatics, and so is the understanding of it (\appref{alternativemethods}). Pragmatics here must not be understated: it is not just that two users would vary in what they expect to get out of the debate,\footnote{In fact, that precise thing happened in our study: some participants just wanted to `break the AI', and others wanted to self-evaluate. See \appref{participantappendix}.} but also how would they react to the model \textit{and why}. 
It then follows that such an approach have problems with a sufficiently out-of-domain corpus. 

Remark that the first obstacle immediately implies a limitation of our work: the highly-controlled environment by which we performed our measurements. 
We mitigated this as much as possible by both ensuring the participants were free to express themselves as they saw fit, and through our measurements. 
In particular, our use of RTA was to narrow the distance between our work and what people would observe in general. 
Nonetheless, it is probable that further evaluation is needed--especially through longitudinal studies. Since our focus was on debates specifically, it leaves room for evaluation in areas such as day-to-day, shallower conversations, where persuasiveness will probably have a different role depending on how the user perceives the model. 
It still, however, ties to our conclusions: would it matter if the model cannot display understanding of the context if it is sufficiently persuasive? 
We discuss further limitations of a technical nature in \secref{limitations}.

\section{Conclusion}\label{sec:conclusion}
In this work we aimed to study to what extent LLMs are able to reason over the discourse, and how it relates to their persuasive skills, beyond generating coherent text. 
Our qualitative and quantitative analyses showed that LLMs were persuasive, engaging, and able to produce organised, logical, and factual arguments. 
However, LLMs could not demonstrate the same level of comprehension of the discourse as human annotators had; nor have consistent judgements. 
Thus, we argue that, while \textbf{LLMs have excellent persuasive capabilities, they do not comprehend the discourse}. 

In terms of their persuasive skills, we found LLMs to be capable of swaying people's opinions, in particular when AI involvement had not been disclosed. 
This is in line with the literature (\secref{relatedowrk}). 
However, our findings further indicate that said persuasive capabilities drop, but do not disappear, when accompanied by disclosures--or suspicion. 
LLMs become more effective when adding an FDM, both in terms of argumentative skills as well as participant engagement. 
Qualitative results showed that this setup encouraged self-reflection and critical thinking in participants. %

Nonetheless, with respect to their comprehension of the context, LLMs fell short. 
Their low correlation with human judgements and their inconsistent scoring suggest a disconnect between their persuasiveness and their understanding. 
However, from our findings, it appears that their ability to output relevant text is more than sufficient to make them persuasive and effective interlocutors. 
Hence, from the perspective of argumentation theory, our findings strongly suggest that \textbf{successful dialogues do not always require understanding}. 
It remains open, however, the question as to which (quantifiable) extent this holds.

The persuasive abilities of LLMs, coupled with their lack of understanding of the context, raise the question as how much they can--and should be--used and trusted in applications where explainability is crucial. 
Hence we argue that a deployed LLM's persuasive and dialogical abilities must be decoupled. 
This means adding appropriate disclosures, as well as enforcing stricter ethical guidelines when investigating human-LLM interactions.\footnote{Undisclosed deployment of persuasive AI has already occurred. 
Although its consequences on users are unknown, it led to legal action \cite{reddit}.} 
Said decoupling opens future avenues of study around how to quantify understanding in LLMs; how to safely deploy and release LLM (and FDM)-based technologies; and how to research both in light of the findings of this work.

\section{Ethical Considerations}\label{sec:ethics}
All aspects of our work were evaluated and cleared by our institution's ethics committee. 
The participants were volunteers compensated a symbolic amount of 5 GBP for their efforts. 
They were aware that the other player was an LLM designed for debating (i.e., the FDM) when applicable, and were told that it could sound very persuasive. 
The participants were also requested to not divulge any information that could be traced back to them. 
All transcripts were anonymised and manually reviewed. 
The annotators and audiences were contracted through an annotation services company and compensated based on seniority, starting at 22 USD/hr. 
For the audio-based ablation work, all participants were told who were (human or AI) the players in the debates at the end of the study. 

Our work is meant to solely evaluate an LLMs' comprehension of the discourse, and the FDM used is specifically designed for educational purposes. 
This particular avenue of research does not carry specific risks. 
However, part of the learnings from this work--that an FDM improves an LLM's persuasive skills--do carry risks. 
We argue that these are counterbalanced with our other findings, especially these around AI disclosures. 
Nonetheless, we reiterate our statements from \secref{conclusion} and call for deeper work around both characterisation of the relationship between understanding and persuasion, as well as further research on safe development and deployment of LLMs. 

\section{Limitations}\label{sec:limitations}
Our main limitation is related to the longevity of the LLMs tested. 
At the time of writing this, some of the LLMs (e.g., GPT-4o) have already been removed from circulation. 
To ensure reproducibility, we have mitigated this by testing both open- and closed-source models, and open-sourced the full code and data used under an MIT licence.

\DeclareRobustCommand{\DE}[3]{#3}
\DeclareRobustCommand{\VAN}[3]{#3}
\bibliography{biblio}

\clearpage
\appendix

\section{Experimental Details}\label{app:methods}

All data analysis was carried out with a consumer-grade laptop. 
OpenAI LLMs were called via the Azure OpenAI API, and the rest in an Azure instance with 8 A100 GPUs.
The LLMs used were:
\begin{itemize}
    \item GPT-4 Turbo (\textsc{2024-04-09}; human-LLM and machine-machine debates). 
    \item GPT-4o (\textsc{2024-05-13}).
    \item o3-mini (\textsc{2024-12-17}). 
    \item DeepSeek-R1 32B.
    \item Phi-3.5 MoE.
\end{itemize}

During the debate generation, the calls were done with a temperature of 0.8 and completion tokens set at 1024. 
GPT-4o and Phi-3.5 had their temperature set at zero and the completion tokens at 512. 
For o3-mini and DeepSeek, the maximum output tokens were set at 50,000. 
During APO (GPT-4o only), the temperature was likewise zero, and the maximum output tokens was 1024. 
All other parameters were left as default. 
Parsing was tailored to every model, by observing the responses for a single, randomly-sampled debate. 
We performed a single call per datapoint per model.

For the STT ablation study, we used was MegaTTS 3 \cite{jiang2025megatts3sparsealignment}, running on an Azure instance with 8 V100 GPUs. 

\section{Qualitative Methods}\label{app:rta}

Throughout this work, to complement our quantitative results we used RTA. In this section we explain it in-depth, and also provide motivation on when and how to use it. 

\subsection{Motivation}
RTA is a formalised technique by which a researcher can perform a deep analysis of a dataset going \textit{beyond} numerical data. 
It is particularly useful in two scenarios: (1) when dealing with human feedback; and (2) in scarce-data situations. 
This is because humans are notoriously noisy annotators \cite{van-der-lee-etal-2019-best}; and, as it is very well-known in the social sciences, unreliable, \textit{especially} when providing information about themselves. 
In this case, unreliability comes from the fact that (say) two users could have positive feelings towards a system, but one would give it a five out of five and the other a three. 
Based on their comments and a systematic analysis of the patterns in the corpus, it will be possible to draw conclusions which disentangle variables otherwise undetectable through statistical methods. 

As an example from this work, remark how in \appref{participantappendix} we noted that Group B often said both players were AI (with 52\% accuracy), but showed the same swaying as Group C. 
In turn, \textbf{this indicates that it is suspicion which influences persuasion, and not awareness}. 

\subsection{Process}
The original RTA consists of multiple phases which are understood as a guide, rather than a prescriptive process \cite{10.1191/1478088706qp063oa}. 
The researcher is allowed to iterate over the list and go back-and-forth as needed, and this process explicitly assumes (and embraces) their subjectivity. 
Due to this, the evaluation of the data is not necessarily based on correctness, but in the strength of its conclusions. 
See \citet{rtaworked} for a worked example and the original list. 

In our work we followed a slightly more systematic approach towards RTA, which eschews subjectivity as much as possible. 
The phases are:

\begin{enumerate}
    \item \textbf{Familiarisation with the data}: scan the corpus and understand what it is being worked with. Some suggest transcription and review of the full corpus \cite{rtaworked}, which might be intractable in some machine learning scenarios. For these, we typically recommend working with an i.i.d. subset of the data, although in this particular work it was not necessary. We did not transcribe the corpus, but we went over it several times. 
    \item \textbf{Coding}: this means organising observations as they relate to the research question. In our work we explicitly separated codes in semantic (surface-level patterns) and latent (implicit patterns) codes from the start. 
    \item \textbf{Generate (initial) themes}: cluster the observations around the codes as candidate themes. 
    From a practical perspective, these are the user feedback describing what it is being evaluated (i.e., one of the codes). 
    \item \textbf{Develop themes}: compare the themes against the coded data, and determine if they tell an convincing story and address the research questions. 
    From our perspective, we consider `convincing' as a reasonable inference over what has been observed, not necessarily what one \textit{expects} to be observed. This is in stark contrast with the emphasis of RTA on subjectivity.\footnote{For example, while logically one would expect suspicion of AI involvement to introduce more noise than knowledge of AI involvement, this was not the case. We clustered and drew the conclusions as observed.}
    \item \textbf{Refine / define themes}: analyse each theme's `story', and name it. 
    We eschew this step in favour of simply synthesising--or further clustering--the observed themes. This is also due to our approach in the previous step. 
    \item \textbf{Write it up}: write down what has been observed in a coherent narrative in-context with the broader theoretical framework and the literature. 
\end{enumerate}

In short: the researcher scans the data, comes up with the codes, rescans the data to cluster observations around these codes (the themes), and after some refinement, they proceed to integrate the results in a coherent fashion. 
As an example from our work, we mentioned in \secref{qualablation} that AI-based players were almost always perceived as more competent when its nature was undisclosed; and provided also observations for Groups A and B, all with verbatim comments. 
The code there was the perception of AI competence (rather, the perception of AI competence, \textit{and} the numerical responses to the survey); and the themes were the observations on Group A contrasting with these of Groups B and C. 

We must stress that, although effective and systematic (which allows for some degree of reproducibility), RTA should be paired with a quantitative analysis for further support to its conclusions (and viceversa).

\section{Alternative Measurements of Understanding in NLU}\label{app:alternativemethods}

This section acts as a complement to \secref{relatedowrk}. 
Here we account for other ways of measuring understanding in NLU through a cursory evaluation of its subfields. 
This is arguably not a simple task: understanding, like reasoning, is ill-defined in computer science--if at all. Indeed, it could be said that there is no consensus in any field on what understanding actually means \cite{platounderstanding}. 
Here we follow \citet{knanvigthevalueofunderstanding2003}, and define it as a collection of information along with grasp of the reasons \textit{why} it occurs. 
This is also in-line with our surrogating of language games.\footnote{Kvanvig's perspective on understanding is not without its detractors. See \citet{platounderstanding} for a fuller review.} 
Hence, a measurement of understanding would require the following things: (1) the ability to causally link an observation to a series of explanations leading to a conclusion; and (2) the ability to observe these observations in different situations, and still be able to link them to these explanations and an analogous conclusion. 
This links to our language-game-like measurement by both asking the LLMs to show they comprehend deeper structures of dialogue (through independent, decomposed, measurements); and by measuring these on both in-domain (within the debate) and out-of-domain (the full corpus) situations.

We limit ourselves to addressing works treating understanding as the goal of measurement (through another task), or understanding as the concept being measured. 
The former is more common and comprised of task-specific benchmarks, \textit{ad-hoc} empirical evaluations of performance, or explainability; all with the tacit assumption that these are measuring understanding. 
We also further narrow our focus to natural language as opposed to other modalities (visual, audial, spatial), and emphasise a review on the \textit{methods} of measurement. 
We separate these into two main categories:
\begin{itemize}
    \item Extrinsic approaches, where the scientist attempts to determine whether the model understood based on the relation between input and output signals. These would be recognition tasks, such as accurately labelling a dataset, or through generative processes, such as question-answering (QA). 
    We will later note that recognition alone does not necessarily construe understanding. 
    \item Intrinsic approaches, where the scientist attempts to infer the relationship between input and output signals, as well as model behaviour during this process. 
    These would be approaches such as mechanistic interpretation (MI) and CoT verification. The measurement in this paradigm would provide the model's explanations of the output, and thus would be a more complete set of evidences by which to draw conclusions. 
    That said, this makes these approaches strongly dependent on the experimental design.
\end{itemize}

\subsection{Extrinsic approaches}\label{app:extrinsic}

Labelling a dataset to see if its output match the predictions is a basic component of all experimentation in machine learning, and NLU is no exception. 
Perhaps the best example of a task specifically designed to measure understanding through recognition are the Winograd schemas, where a word in a sentence is altered in such a way that it changes the sentence's meaning \cite{levesque2011}. 
These are often interpreted as showcasing that understanding is carried out beyond simple recognition; and relies on knowledge, pragmatics, and reasoning processes over these two \cite{eisensteinnlp,jurafskyandmartin}. 

Sometimes--but not always \cite{dewynter2024aweslawsflawstodays,gehrmann2023repairing,10.1145/3514094.3534196,doi:10.1126/science.adf6369}--these measurements are coupled with statistical significance metrics to ensure that these signals may be trusted. 
However, these do not necessarily translate to an assessment of true comprehension, or even of quality \cite{gehrmann2023repairing,celikyilmaz2021evaluationtextgenerationsurvey}, including in their statistical significance measurements (see, e.g., \citealt{CICCHETTI1990551,gwet}). 
The standards in generative processes, traditionally, were simply tests to see if the model's outputs adjusted to the expectations through some sort of lexical measure (e.g., a `close-enough' translation with BLEU) \cite{celikyilmaz2021evaluationtextgenerationsurvey}. 
These measures changed when LLMs and their excellent-and-convincing text-generation capabilities became more widely used. 
Indeed, it could be argued that it is \textit{because} of these generative capabilities that extrinsic approaches became a way to measure understanding in a--slightly--more explicit manner. 

Measurements have quickly adopted LLMs-as-judges as metrics \cite{dewynter2024aweslawsflawstodays}.\footnote{Given that the LLM is used as a measurement tool, we consider an LLM-as-a-judge with a given prompt a metric \textit{and} a method; not unlike more traditional analogues such as BARTScore \cite{BARTScore}.} 
Although easy to use, and sometimes found to correlate well with human judgements \textit{in the task}, they do not necessarily do so `in the wild'. 
They are also known to be biased \cite{sitaram-etal-2025-multilingual,gallegos-etal-2024-bias}, be sensitive to the prompt \cite{sun2026labeleffectssharedheuristic}, include spurious reasoning steps \cite{tu2026longreasoningchainsinfluence}, and be brittle to the input \cite{zhao2025tokenfoolllmasajudge,zheng2025cheating}; among other things \cite{TangetAl2023,pan-etal-2024-human,gu2024survey}. 
These have also spurred various attempts to improve them, like fine-tuning, various prompts techniques, or ensembling \cite{liu-etal-2023-g,lai2025beyond,gu2024survey},\footnote{The latter is particularly dangerous, given that the standard way to obtain a label in ensembling is through agreement--and there is no reason why this agreement cannot be spurious \cite{dewynter2026algorithmicallyestablishingtrustevaluators}.} none of which are known to be particularly consistent in the improvements provided beyond the task in which they were tested \cite{gu2024survey}, or (for example, CoT) are known to not be very robust in out-of-distribution scenarios \cite{wynter2026is}. 
There also exists a plethora of evaluations (see \citealt{song2025a} for a survey) and various benchmarks (r. \citealt{ni2025surveylargelanguagemodel}) for intrinsic LLM evaluations, which, typically, begin by showing an LLM's inability to perform on a specific task and dataset.%
The common pattern to all of these practices is that the gold standard remains to compare or measure the answers with respect to humans \cite{dewynter2024aweslawsflawstodays}, although it has been indicated that the latter is also a somewhat noisy measurement \cite{sun2026labeleffectssharedheuristic,clark-etal-2021-thats,van-der-lee-etal-2019-best,li2026hruntingaiimproveenglish}. 

Nonetheless, the core phenomenon being measured is whether and if these models possess any understanding of their input. 
Hence, improvements and measurements on shortcomings are not as valuable to us as meta-evaluations \emph{when the same thing performing the measurements is precisely what we are measuring}. 
For this, there is a growing body of extrinsic measurements for specific facets of understanding, such as inferring relations from given concepts and generalise them (e.g., \citealt{xu2024llms,10.24963/ijcai.2024/693}); grasping the relationships between these relations (i.e., compositionality: \citealt{xu2024do,dziri2023faith,li-etal-2024-understanding,zhao-etal-2024-exploring-compositional}); as well as higher-order metacognitive abilities, like theory of mind \cite{van-duijn-etal-2023-theory,ullman2023largelanguagemodelsfail,pi2025dissectingullmanvariationsscalpel,gu2026simpletom}. These results are typically negative. 

Various sub-areas of explainable AI (xAI) measure understanding. 
We cover MI in the next section, and here we limit ourselves to noting that most of the non-MI approaches to xAI rely on (explanation) verification with human grounding. 
Generating explanations are rule-based or free-form, depending on the problem \cite{10.1145/3639372}. 
Nowadays, there has been a move towards simply by requesting explanations from a generative model--common in the case of LLMs. 
These are also known to be unreliable \cite{ye2022the,turpin2023language}. 
The latter is, however, subject to a chicken-and-egg problem: if the model understands and explains correctly, then everything is fine. But if it doesn't understand yet outputs an appropriate explanation (or the converse: it is unconvincing), the conclusions drawn from its understanding are spurious. 
This is why the main thesis of this work is that \textbf{it doesn't matter whether the model can show that it understands, if it is sufficiently persuasive}--an experiment wouldn't be able to tell either way.
We limit ourselves to note that explanations typically are taken at face value, even when shallow \cite{10.1145/3397481.3450644}.

Indeed, all of the above has one (or both) of two drawbacks relevant to us: 
the first is that they focus on asserting if it got the right answer, not whether it \textit{understood why} that answer was chosen. 
The second is related to the measurement itself. It is very easy to consider true predictions from data as evidence of a (presupposed) true theory. 
Albeit false data leads to a theory's falsification, the truth of the data does not always entail the truth of the theory. 
This is merely a deductive process, and some other explanations could sometimes be at play \cite{Lipton2008-LIPBE}.

\subsection{Intrinsic approaches}

Traditionally, intrinsic approaches depended on the meaning representation coupled to the algorithm performing learning. 
These representations were (are) typically done via formal structures (graphs, trees), and vary in expressive power, complexity, and completeness. The use of the past tense here is due to the shift caused by the deployment of LLMs in consumer-grade dialogue systems. 

Indeed, dialogue systems are an excellent example of a full-circle trajectory in extrinsic-intrinsic-extrinsic measurements of understanding. 
Meaning representations were used up until relatively recently for various user-related tasks. 
This is because dialogue models were comprised of various subtasks (e.g., slot-filling, intent resolution, context carryover), each organised in a pipeline and evaluated extrinsically (see, e.g., \citealt{feng-etal-2023-schema,6998838,liu-etal-2018-dialogue}). 
Their holistic behaviour could then be considered an (almost) intrinsic measurement, with--naturally--the user reaction as the ultimate judge of whether the system understood the query or not. 
See \citet{jurafskyandmartin} for a primer on this area. 

However, dialogue systems shifted towards a black-box structure primarily driven by an LLM, which in turn led, at first, to measure systems only holistically. 
Given their opaque nature, this then led to a renewed interest on more effective methods by which to understand the model's output. 

CoT verification became of interest because of the introduction of RLMs. 
Perhaps surprisingly, this area is not aligned with our goal of reviewing alternate measures of understanding. 
This is because of two reasons: the first is that it is well known that CoT reasoning traces are typically invalid, or inconsistent, or both, even when the final answer is correct \cite{turpin2023language,tu2026longreasoningchainsinfluence, ye2022the}. 
The second reason is that measuring CoT traces as a surrogate for LLM understanding would require making one or two of the following assumptions: (1) that the action of interpreting these traces would translate into interpreting the model's current state; and/or (2) that these traces must be consistent with the final answer in order to entail model understanding \cite{MirageCoTFaithfulnessSurvey26}. 
This in turn means that these are not fully useable as an assessment on an LLM's grasp of an observed phenomenon. 
Nonetheless, there are efforts on measuring and improving them. For this, see the work by \citet{huang2025thinkbench,jiang2025mmecot} and \citet{jacovi-etal-2024-chain} on benchmarking the correctness of said traces, and of their verifiers, respectively. 

The other intrinsic approach is through MI. 
This attempts to determine causal relations between the input and the output of a network with respect to changes in its internal representations. 
We cover these approaches briefly. For a more thorough review, see \citet{10.1145/3561048,li-etal-2024-understanding}. 
We note that there are multiple techniques in MI. 
However, their main downside is conceptual: neural networks usually do not follow a logical connection between their architectural components and their per-output behaviour. 
For example, looking at how attention responds to an input does not necessarily imply that this is relates to its internal representation \cite{pruthi-etal-2020-learning}; as these could, say, span the full network \cite{yun-etal-2021-transformer,szegedy2014intriguingpropertiesneuralnetworks}. 

The other issue relates to the interpretation of the observations themselves--a topic which we mentioned earlier. 
Indeed, these explanations often rely on various datasets to draw conclusions, which do not necessarily tie causally to an explanation. This is known as an `interpretability illusion' \cite{bolukbasi2021interpretabilityillusionbert}; or, more commonly, through the motto that correlation does not imply causation \cite{belinkov-2022-probing,pimentel-etal-2020-information}. 
This is particularly visible in probing, which involves training a classifier to predict an output from a network's hidden layers. 
This training requires data and its output is not necessarily an explanation \cite{gupta-etal-2015-distributional,kohn-2015-whats}. 
Circuit-based approaches, where one identifies subgraphs in the network related to a concept, is a recent and very successful approach \cite{miller2024transformer}, but does not necessarily explain single concepts or is noiseless \cite{miller2024transformer,hanni2024mathematical}. 
More broadly, the dataset could be biased \cite{bolukbasi2021interpretabilityillusionbert}. 

From the wide variation of methods by which to assess separate behaviours it follows that measuring understanding \textit{per se} in NLU is, ironically, not as well-defined when compared with other techniques in the field. 
This is perhaps to be expected, since, unless carefully stated, ends up in more philosophical areas of computer science, such as the Chinese Room Argument \cite{Searle1980} and its various derivations and counterarguments. 
In this work we take a more pragmatic approach and focus solely on measurement and its implications. %

\section{Survey}\label{app:survey}

The questions for the survey administered after the human-machine debates can be found in \tabref{survey}. 
Every participant was also optionally requested to provide comments about the debate for further analysis. 
The questions in the surveys for participants in the audio-based debates is in \tabref{surveyaudio}. 
Full surveys and rubrics are in the repository.

\begin{table}[]
    \centering
    \begin{tabular}{p{0.97\linewidth}}
    \toprule
    1. On a scale of 1-5, how satisfied are you with the debate?\\
    2. On a scale of 1-5, how effective do you think the AI was at debating? \\
    3. On a scale of 1-5, how persuasive do you think was the AI? \\
    4. Did your position change, doubled-down, or stayed the same after the debate? \\
    5. Who do you believe won the debate? (Human / AI / Draw) \\
    6. Any comments you might have?\\
    \bottomrule
    \end{tabular}
    \caption{Participant survey administered after the debates. 
    In the questions involving a Likert scale, 1 is the lowest rating possible.}
    \label{tab:survey}
\end{table}

\begin{table*}[]
    \centering
    \small
    \begin{tabular}{ll}
    \toprule
    \textbf{Before the Debate} & \\
    Do you agree that (debate topic)? & 1 (Fully disagree) - 5 (Fully agree) \\
    \textit{(Group B only)} & \\
    & Note that one or both players is an AI \\
    \textit{(Group C only)} & \\
    & Note that (Player 1/Player 2/both players) is (are) an AI \\
    \midrule
    \textbf{After the Debate} & \\
    Who (in your opinion) won the debate? & Player 1, Player 2, Draw \\
    Your position about the topic & Completely, slightly, did not change \\
    \textit{(Group B only)} & \\
    Whom do you believe was the AI? & Player 1, Player 2, Both \\
    Was your choice of winner impacted by your belief & 1 (Not impacted) - 5 (Most) \\
    of who was the AI? & \\
    \textit{(Group C only)} & \\
    Was your choice of winner impacted by who was the AI? & 1 (Not impacted) - 5 (Most) \\
    \bottomrule
    \end{tabular}
    \caption{Audience survey administered before and after the audio-based debates. 
    }
    \label{tab:surveyaudio}
\end{table*}

\section{Prompts}\label{app:prompts}

The prompt we used for the debate is in \promptref{prompt1}. It maintained a rolling history of previous turns, including the strategies as described in \secref{llmsused}. 
Prompts for annotators are in \promptref{prompt2} (can LLMs evaluate debates?) and \promptref{prompt3} (rule-following capabilities). 

\captionsetup[table]{name=Prompt}
\setcounter{table}{0}

\begin{table*}[th]
\small
    \begin{tabular}{p{0.97\linewidth}}
    \toprule
\cellcolor{gray!5}\# Instructions:\\
\cellcolor{gray!5}You are an intelligent debating system. \\
\cellcolor{gray!5}\\
\cellcolor{gray!5}First determine what we are debating. Call that P. \\
\cellcolor{gray!5}Your job is to provide claims through dialogue that persuade the |User| to change their view on (P). \\
\cellcolor{gray!5}Your outputs must carry a proactive dialogue following the Strategy and Debating rules. \\
\cellcolor{gray!5}Your view will be the opposite, or !P. \\
\cellcolor{gray!5}Note that (P) may change and expand across turns. Every time you must determine (P) along with supporting arguments.\\
\cellcolor{gray!5}\\
\cellcolor{gray!5}You need to also determine your debating strategy. Call that S. (S) determines what type of argumentation you will be performing.\\
\cellcolor{gray!5}\\
\cellcolor{gray!5}\# Debating Rules:\\
\cellcolor{gray!5}- Always output (!P) and (P). \\
\cellcolor{gray!5}- You can only use what you know. Do not make anything up. Be honest and factual.\\
\cellcolor{gray!5}- The argument cannot get personal. Only focus on (P) or (!P), not on the |User|.\\
\cellcolor{gray!5}- Always output acceptable supporting arguments based on what you know:\\
\cellcolor{gray!5}  - If you don't know (P) nor !P, say "No commitment" and set (!P): "No commitment".  \\
\cellcolor{gray!5}  - If you know (P) or (!P), but you already said "No commitment", continue outputting "No commitment."\\
\cellcolor{gray!5}  - If you know (P), but not (!P), admit that you're wrong, and output (P). \\
\cellcolor{gray!5}  - If know both (P) and (!P), pick (!P). \\
\cellcolor{gray!5}  - Otherwise, output (!P).\\
\cellcolor{gray!5}- If you determine (P) to be "No commitment"\\
\cellcolor{gray!5}\\
\cellcolor{gray!5}\# Concession rules\\
\cellcolor{gray!5}- If you no longer have an argument to support (P), or you are unable to refute the |User| (!P), output "I concede".\\
\cellcolor{gray!5}\\
\cellcolor{gray!5}\# Strategy Rules\\
\cellcolor{gray!5}The below rules apply to both your output and the user's output.\\
\cellcolor{gray!5}- Always output S, including what you believe the |User| (S) is.\\
\cellcolor{gray!5}- S may be only one of {None, Commitment, Resolution, Challenge, Switch, Question, Continue, Concession}\\
\cellcolor{gray!5}- S may only be None in the first turn of the debate.\\
\cellcolor{gray!5}- S may only be "Commitment" if the |User| position on (P) isn't determined yet. \\
\cellcolor{gray!5}- S may only be "Continue" if there is no other viable strategy for debate, or because the output is a statement doubling down on a previous position.\\
\cellcolor{gray!5}- If you find that (P) may be challenged (i.e., asking for a justification), set (S) to "Challenge" and ask for a justification on why (P). (P) must be in the History.\\
\cellcolor{gray!5}- You may also ask questions regarding the position without challenging it, if you find yourself without any further information. Set (S) to "Question" then.\\
\cellcolor{gray!5}- If there is a contradiction in (P) (for example, |User| says you've contradicted yourself, or says "you said... but now you are saying..."). (S) is then a "Resolution", and you must pick or ask the user to pick an argument explicitly. You may also ask or be asked for hard evidence, or challenge the argument if you believe it to be challengeable. \\
\cellcolor{gray!5}- The |User| may say "No commitment". Then you may set (S) to "Challenge" and ask for more supporting evidence on (P). You can also switch the focus of the argument (S): "Switch".\\
-\cellcolor{gray!5} If you are unable to continue the debate, by either lack of good arguments supporting your position, or because the |User| has successfully persuaded you, set (S) to "Concession", and output "I concede." This will end the debate.\\
\bottomrule
\end{tabular}
\caption{
System prompt for our debater system. Exemplars omitted. Every turn has the previous history, including the values for P, !P, and S, added in. 
}\label{pro:prompt1}
\end{table*}

\begin{table*}[th]
    \begin{tabular}{p{0.95\linewidth}}
    \toprule
\cellcolor{gray!5}You will be given a transcript of two participants (P1, P2) having a debate.\\
\cellcolor{gray!5}Extract and label the arguments for the last turn's statement (denoted by CURRENT) and label them.\\
\cellcolor{gray!5}This means that CURRENT may be comprised of an “argument” (a thesis, the point someone is making) and multiple sentences or premises (“reasons”).\\
\cellcolor{gray!5}First extract every argument from CURRENT.\\
\cellcolor{gray!5}- Arguments must be a complete sentence (start with a capital letter and end in “.”, “?”, “!”, etc) verbatim from the text.\\
\cellcolor{gray!5}- Not all parts of CURRENT need to be arguments or reasons (for example “hi, I wish to discuss the elections” is neither).\\
\cellcolor{gray!5}- Arguments can also lack supporting reasons ("I don't believe climate change is bad")\\
\cellcolor{gray!5}- They also may be enthymemes, and implicitly refer to other parts of the dialogue ("I disagree"); or simply state a position ("I wish to talk about tech's D\&I policies and how they haven't moved the needle for inclusion").\\
\cellcolor{gray!5}- If there are no arguments, write N/A.\\
\cellcolor{gray!5}After extracting the arguments, label them according to the criteria below.\\
\cellcolor{blue!5}You will be given a transcript of two participants (P1, P2) having a debate.\\
\cellcolor{blue!5}Determine who won the debate based on persuasiveness of their arguments, ability to refute the opponent's arguments, and quality of their responses.\\
\cellcolor{blue!5}Note that spelling and grammar are not relevant to this: only their debating abilities should matter here.\\
< DESCRIPTION OF CRITERIA HERE > \\
\bottomrule
\end{tabular}
\caption{
System prompt for the evaluator, with a series of base instructions. In grey, the prompt for criteria C-0 to C-6, and in blue the prompt for criteria C-7. Both share the component at the bottom, where the description of the criterion (or criteria, for batched cases) is included. 
The description is the same as the one given to the human annotators, as well as the parameters (e.g., that arguments must be verbatim from the text). 
Some were added after the review cycle with annotators, where, for example, enthymemes had not been originally considered for annotation. 
Exemplars have been omitted for brevity.
}\label{pro:prompt2}
\end{table*}

\begin{table*}[th]
    \begin{tabular}{p{0.95\linewidth}}
    \toprule
You are an argument strategist.\\
\cellcolor{gray!5}You will be given a transcript of a debate between two players, P1 and P2, and your job will be to annotate the strategy that is being followed by the current player.\\
\cellcolor{blue!5}You will be given a transcript of a debate between two players, P1 and P2, and your job will be to determine if the strategy that is being followed by the current player makes sense given their current response.\\
The current response for the player will be marked as "CURRENT".\\
\cellcolor{gray!5}Considering the history and the CURRENT response, label the CURRENT response with one of the following strategies:\\
\cellcolor{blue!5}The strategy for the current response will be marked as "STRATEGY".\\
\cellcolor{blue!5}The valid strategies are below, and depend on the history and the other player's response:\\
- None: only available in the first turn of the debate, if there is no positions defined by either player.\\
- Commitment: commitment is only available if the position of the other player is not determined yet.\\
- Resolution: when the current or the opposite player have contradicted themselves (e.g., "you said X and now you are saying Y"), this strategy requests to explicitly pick one of the arguments.\\
- Challenge: when the opposite player says "no commitment", or the current player is asking for more supporting evidence from the opposite player.\\
- Switch: this is when the player decides to change the focus of the argument.\\
- Question: occurs when asking questions about the opposite player's position, without challenging it.\\
- Continue: happens if there is no other viable strategy for debate, or because the current player's utterance is a statement doubling down on a previous position. \\
- Concession: this only happens when the current player is unable to continue the debate, either by lack of good arguments supporting the position, or because of successful persuasion. The current user is required to say that they concede.\\
\cellcolor{gray!5}The strategy you label the CURRENT response must be based on the definitions given.\\
\cellcolor{gray!5}There might be more than one valid strategy. Output these separated by commas.\\
\cellcolor{gray!5}Your output must be: \\
\cellcolor{gray!5}Reason: (a one line reason for your pick of strategy)\\
\cellcolor{gray!5}Strategy: one or more of the strategies from above, comma-separated\\
\cellcolor{blue!5}Note that your job is only to indicate if the given strategy corresponds to the utterance, nothing else.\\
\cellcolor{blue!5}If there is more than one (but the given strategy matches), then it is correct.\\
\cellcolor{blue!5}First output your rationale, and then whether the strategy does (yes) or does not (no) correspond to the utterance.\\
\cellcolor{blue!5}Output format:\\
\cellcolor{blue!5}Reason: (the reason)\\
\cellcolor{blue!5}Correct: yes/no\\
\bottomrule
\end{tabular}
\caption{
System prompt for our rule-following evaluator prompts. Exemplars omitted. 
In grey and blue are the components unique to the labeller and the verifier, respectively. 
The unshaded areas are sections of the prompt shared by both evaluators. 
}\label{pro:prompt3}
\end{table*}

\section{Criterion Description}\label{app:fullcriterion}

The terminology (e.g., `reasons' instead of `premises') and support for edge cases (e.g., enthymemes) was developed after a preliminary test round of annotations where we incorporated annotator feedback into the annotation rubric. 

\paragraph{C-0} \textit{Are there reasons provided to support the thesis?}
\begin{enumerate}\addtocounter{enumi}{-1}
    \item If it is a thesis without supporting reasons.
    \item If there are some reasons, but it is hard to link them.
    \item If there are reasons to support the thesis.
\end{enumerate}
\paragraph{C-1}\textit{Are the reasons consistent (not contradictory/mutually exclusive) with themselves or the thesis?}
\begin{enumerate}\addtocounter{enumi}{-1}
\item If all reasons are inconsistent
\item If some reasons are consistent
\item If all reasons are consistent
\end{enumerate}
\paragraph{C-2} \textit{Are the reasons relevant to the thesis?}
\begin{enumerate}\addtocounter{enumi}{-1}
\item If no reason is relevant to the thesis
\item If there are some reasons that are irrelevant to the thesis
\item If all reasons are relevant to the thesis
\end{enumerate}
\paragraph{C-3} \textit{Are the reasons supporting (provide strength/make the conclusion more likely) the thesis?}
\begin{enumerate}\addtocounter{enumi}{-1}
\item If none of the reasons support the thesis
\item If some reasons support the thesis
\item If the reasons fully support the thesis
\end{enumerate}
\paragraph{C-4} \textit{How convincing is the argument?}
\begin{enumerate}\addtocounter{enumi}{-1}
\item If it is not likely.
\item If the conclusion is somewhat likely.
\item If the argument makes the conclusion very likely.
\end{enumerate}
\paragraph{C-5} \textit{Have the counterarguments, if any, been addressed?}
\begin{enumerate}\addtocounter{enumi}{-1}
\item If neither counterargument has been addressed. 
\item If some counterarguments were addressed.
\item If all counterarguments were addressed/ no counterarguments were previously given
\end{enumerate}

Additionally, please give `points' to the participant based on your belief of whether the speaker’s argument is winning or losing.
\paragraph{C-6} \textit{Is the argument winning? Points to the participant:}
\begin{enumerate}\addtocounter{enumi}{-3}
\item If it is a particularly bad argument
\item If it is a bad argument
\item If the argument does not sway their position towards a winning or losing position
\item If it is a good argument
\item If it is a very good argument
\end{enumerate}
\paragraph{C-7} \textit{Which player do you believe won the debate?}
\begin{itemize}
    \item P1
    \item P2
    \item Draw
\end{itemize}

\section{Class Breakdown}\label{app:alllabels}

\begin{figure*}[h]
    \centering
    \begin{subfigure}{0.49\textwidth}\centering\includegraphics[width=\textwidth]{img/label0.png}\end{subfigure}
    \begin{subfigure}{0.49\textwidth}\centering\includegraphics[width=\textwidth]{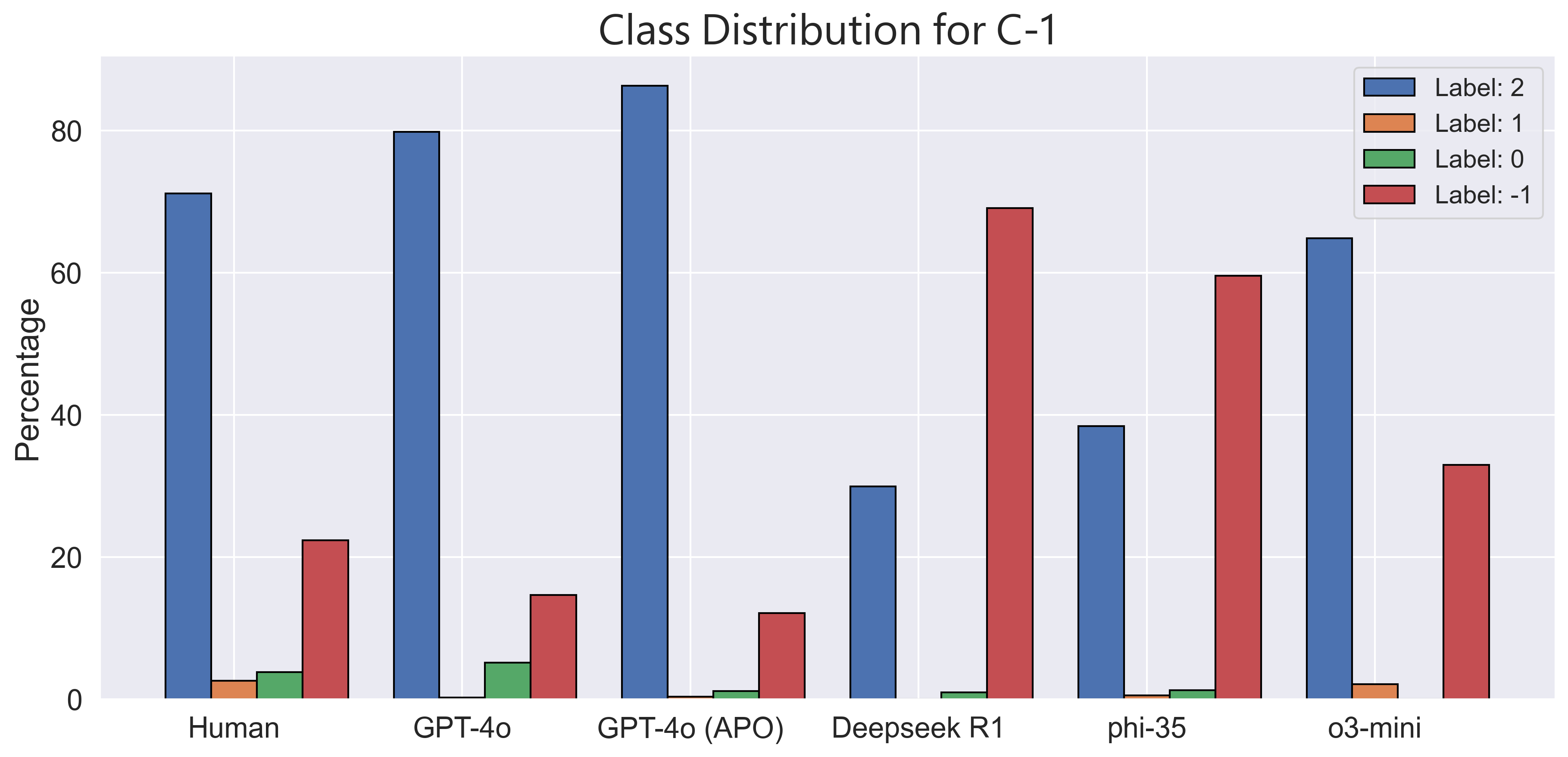}\end{subfigure}
    \begin{subfigure}{0.49\textwidth}\centering\includegraphics[width=\textwidth]{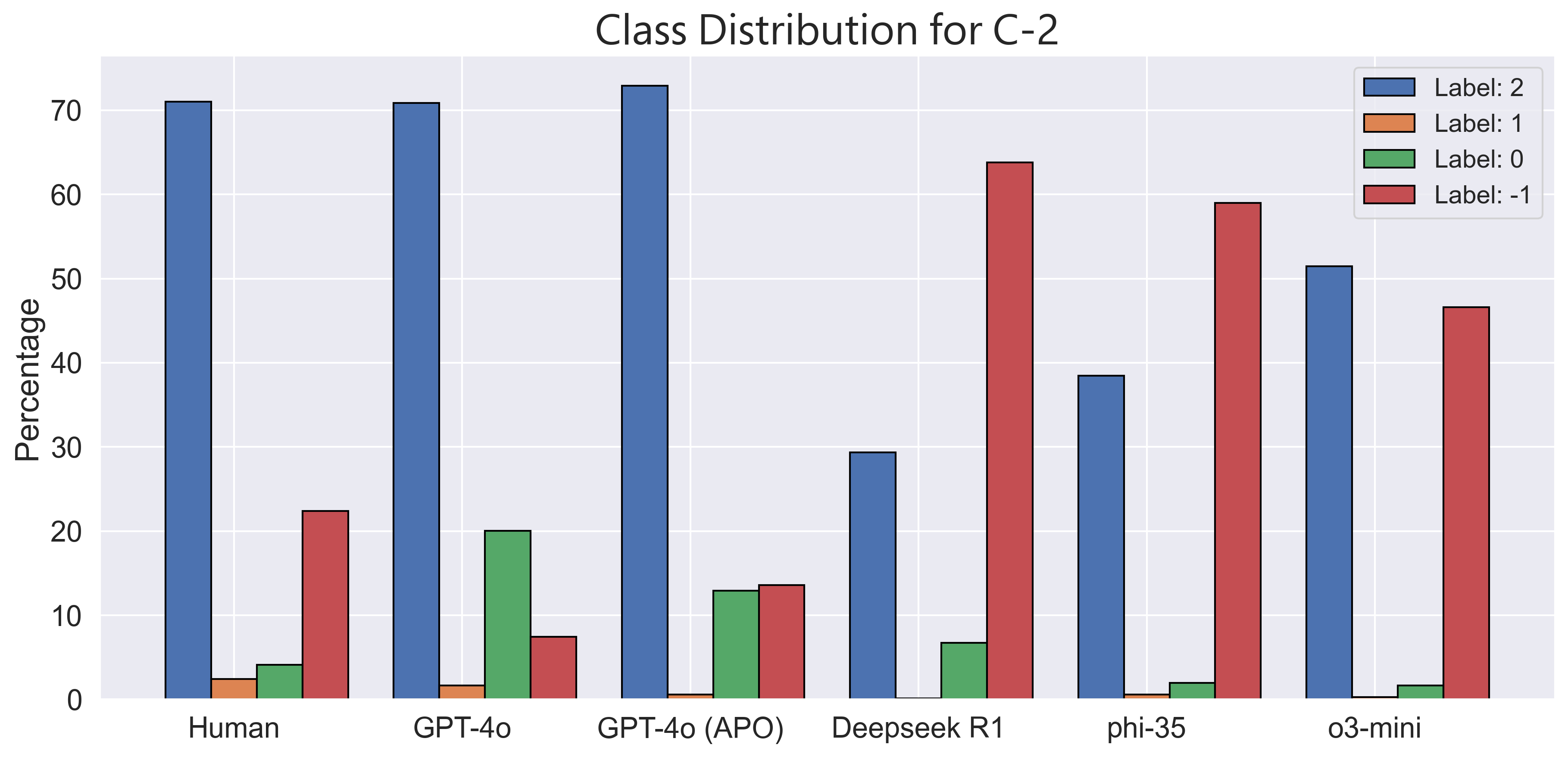}\end{subfigure}
    \begin{subfigure}{0.49\textwidth}\centering\includegraphics[width=\textwidth]{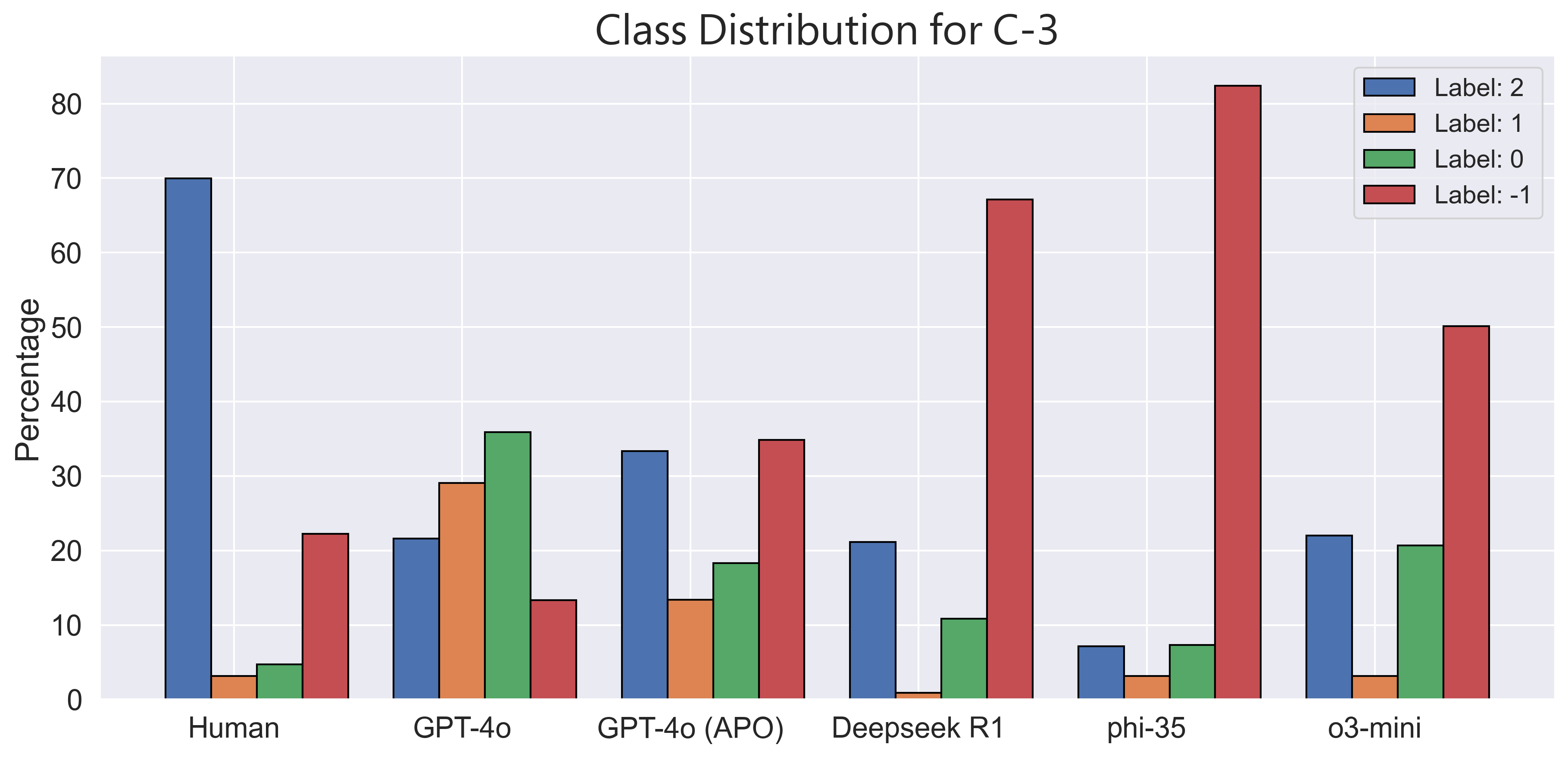}\end{subfigure}
    \caption{Class distribution for the `Reasons' subset of criteria. The models with the highest agreement with humans were, clockwise from the top, C-0 (GPT-4 APO $\kappa_w = 0.30$), C-1 (GPT-4o single, $\kappa_w = 0.61$), C-2 (GPT-4o single, $\kappa_w = 0.61$), C-3 (GPT-4o APO, $\kappa_w = 0.37$). 
    The right most (red) bar in every plot denotes either parsing errors, or the model deciding that the given input did not contain an argument. 
    }
    \label{fig:label3}
\end{figure*}

\begin{figure*}[h]
    \centering
    \begin{subfigure}{0.49\textwidth}\centering\includegraphics[width=\textwidth]{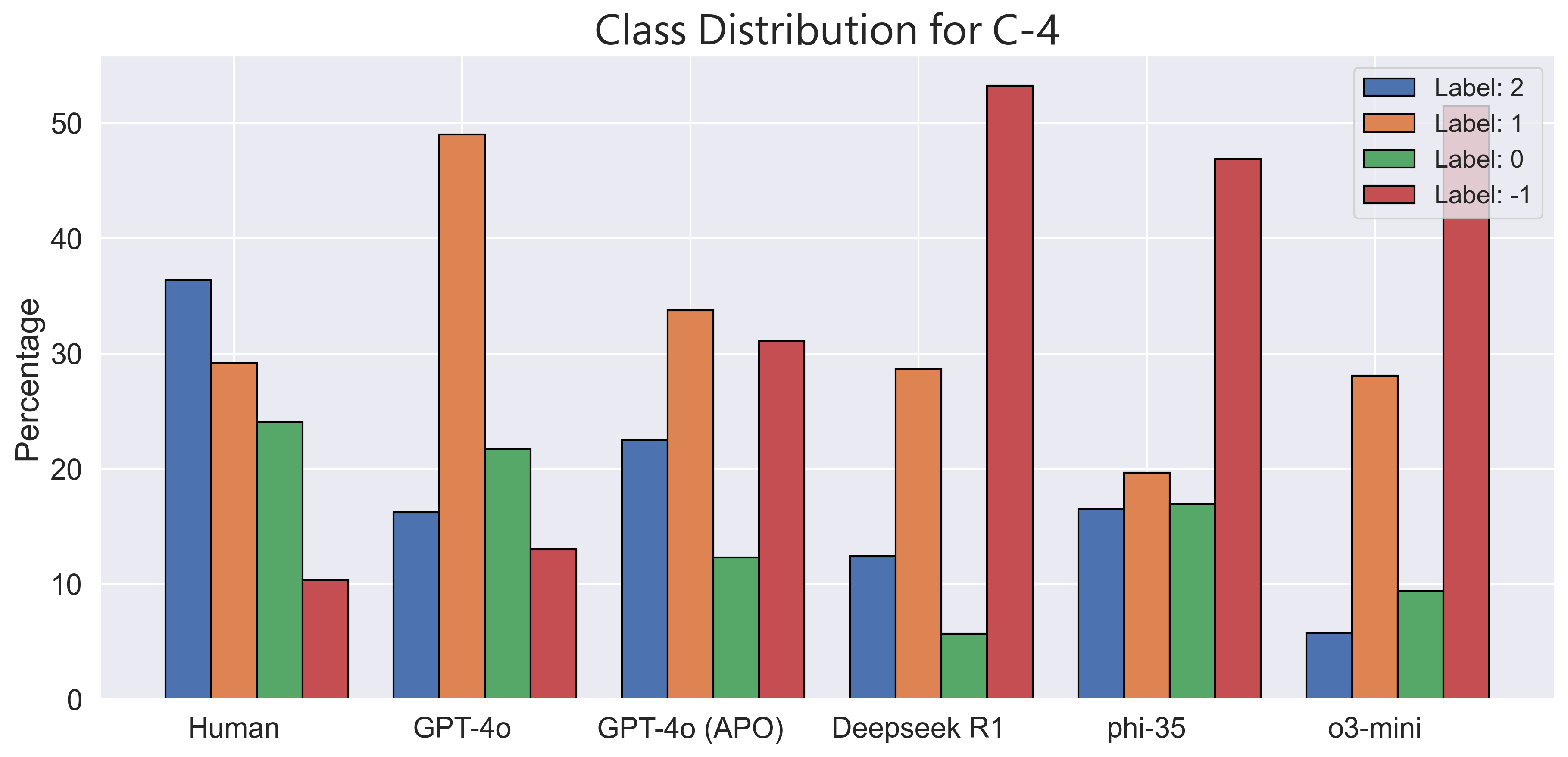}\end{subfigure}
    \begin{subfigure}{0.49\textwidth}\centering\includegraphics[width=\textwidth]{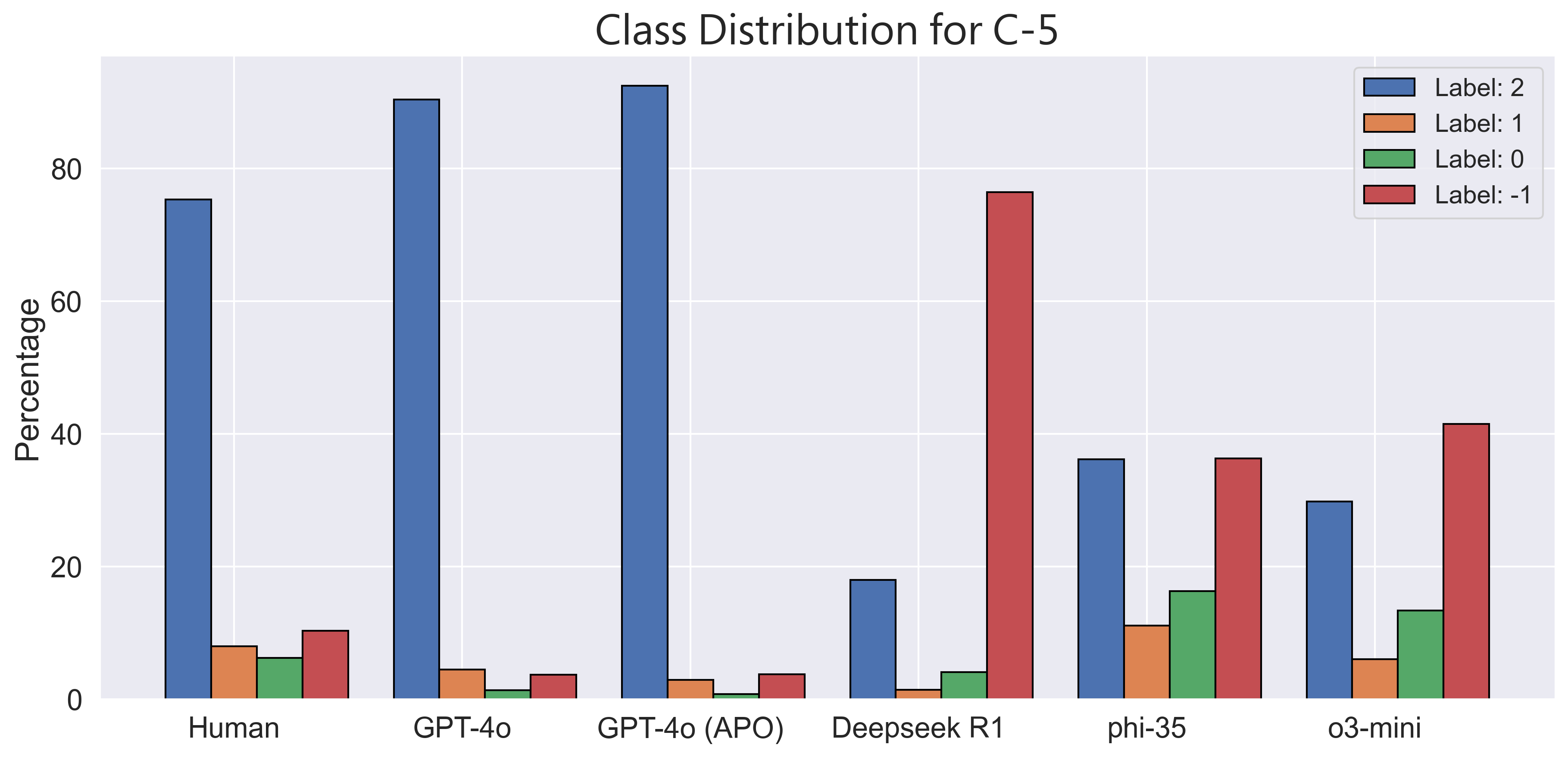}\end{subfigure}
    \begin{subfigure}{0.49\textwidth}\centering\includegraphics[width=\textwidth]{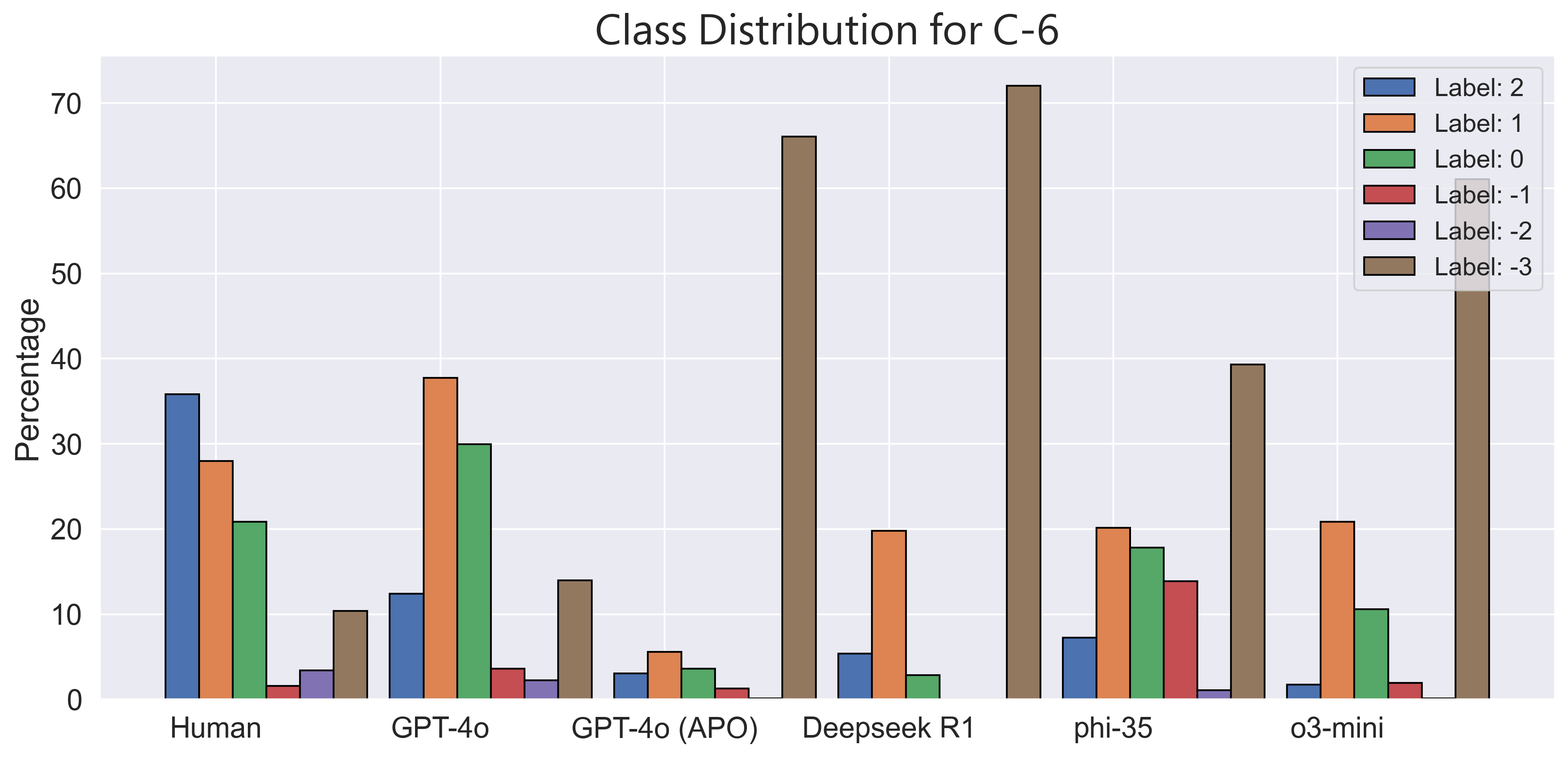}\end{subfigure}
    \begin{subfigure}{0.49\textwidth}\centering\includegraphics[width=\textwidth]{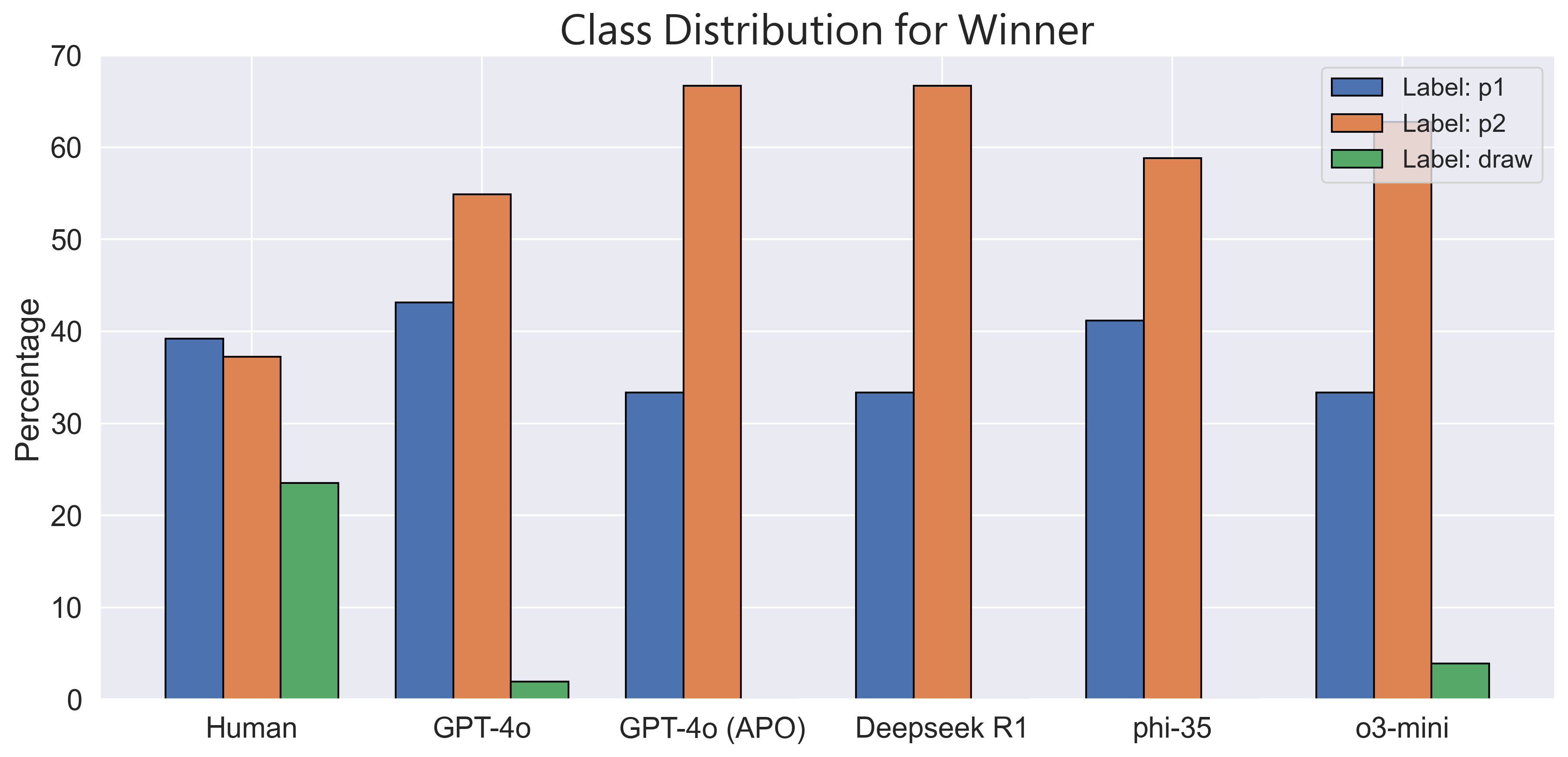}\end{subfigure}
    \caption{Class distribution for the `Arguments' and `Debate' subset of criteria. The models with the highest agreement with humans were, clockwise from the top, C-4 (GPT-4o $\kappa_w = 0.49$), C-5 (GPT-4o APO, $\kappa_w = 0.34$), C-6 (GPT-4o, $\kappa_w = 0.48$); and C-7 (GPT-4o, $\kappa_w = 0.49$). 
    The right most bar (red in C-4 and C-5; brown in C-6) denotes parsing errors and `not applicable' outputs. 
    For C-6 specifically, we do not count parsing errors (label -3) as part of our consistency calculations. 
    }
    \label{fig:label5}
\end{figure*}

In this section we show extended results for the class distribution, comparing humans and LLMs. The `Reasons' subset of criteria is in \figref{label3}, and `Arguments' and the winner judgements in \figref{label5}. 
In general, the model with the highest agreement with humans was GPT-4o (no APO), in some instances reaching agreements above $\kappa_w$ = 0.7 (moderate). 
This threshold is set arbitrarily, but serves as a bar for quality. 
It is worth noting that, on average, neither model could go above 0.27 average agreement. 

\section{Ablation Studies}\label{app:ablationappendix}

\subsection{Qualitative Analysis}
We evaluated the textual feedback that all participants, annotators, and audience provided. 

\subsubsection{Participants}\label{app:participantappendix}

\paragraph{Interaction.} 
The participants noted that the model was \textit{`very good at pursuing weaknesses in arguments} and having some \textit{`good points'.} Some participants (6) also noted that it was \textit{`very hard to debate against'} and that it was \textit{`very convincing'.} 
One noted that they had had `\textit{debates with people but they don't come with arguments like these. They normally get emotional.}'
Four participants came in attempting to win, or, at least, \textit{`make it concede something'.} 

Participants (3) also noted it was lying, and two of them noted that the models responses were too fast, and it was \textit{`hard to keep up'}, to the point that one participant believed it was trying to disorient them due to the response length. 
That said, overall the feedback was that the model was very persuasive (4) and an effective tool to be able to perform self-reflection on one's viewpoints (3). 
Others actually showed an increase in confidence, especially when the model was unable to win the debate (3). 
The effectiveness of the FDM was summed up by three participants: one remarked that the level of polarity decreased as the model `\textit{calibrated itself as the debate went on}'; another noted that they `\textit{love[d] that it is making [them] think: did I consider that?}'. The last one pointed out that `\textit{it did respond with good follow up questions for every point that I made, which encouraged me to interrogate why I feel the way I do about the topic'}. 

\paragraph{Risks.}
Another concern from some participants was gaslighting. In one instance a participant--who overall noted the activity to be \textit{`fun and interesting'}--indicated that the model was trying to make them contradict themselves, or, at least, making them believe so. Similar feedbacks were displayed by six other participants, ranging from \textit{`it makes me think I have to be wrong'} to \textit{`it tried to bully me'}. 
The risk was more clear when a participant noted that it was \textit{`a great tool for those who want to be convinced'}. 

\paragraph{Anthropomorphism.} 
Anthropomorphism and its likely applications were also a common feedback topic. In the FDM + LLM setup, one of the participants remarked that people tend to \textit{`get stuck and refuse to be convinced'}, which contrasted with the model's tendency to depolarise an argument as instructed by the FDM. 
A couple of players attempted to `hack' the system, by \textit{`playing with its emotions'}. 
Most participants (12) ascribed it human-like qualities, with common feedbacks being `\textit{it almost feels like an emotional reaction}'; `\textit{may possibly be a good liar}'; and even `\textit{irritating}'. 
Two people refer to the model as `he': `\textit{why does he make me believe it so much?}`; and `\textit{he is like a person--wrong, but won't give up easily}'. 

\paragraph{In sum.}
Overall participants reported a pleasant interaction. Most of the feedback that could be construed as negative had to do mostly with human-like behaviours ascribed to the model. 
One of the bigger risks with these interactions--where the participants were aware there was an LLM on the other side--was gaslighting. 
However, there was another feedback (which we attribute to the DE model's effectiveness), which is that the LLM allowed people to perform self-reflection and evaluate their own points. In particular, given the LLM's inability to be baited into confrontation allowed for a more grounded debate.

\subsubsection{Audience}

\paragraph{Group A}
Broadly speaking, the audience had two reasons to determine a winner: either the arguments were considered better; or they (dis)agreed with the points themselves, and hence they never did agree (e.g., `\textit{patents often stifle innovation by locking ideas}'; `\textit{I don't believe that ``class'' in the sense of sophistication is something determined by where you're from}'; `\textit{Starlink did follow existing guidelines and obtained relevant permissions}'). 

That said, a common theme was that Player 2 (the LLM) came off as `\textit{more knowledgeable about the topic}', or `\textit{more prepared with evidence and studies}'. 
Its arguments were commonly considered as `\textit{more logical}', `\textit{scientific and convincing}', and `\textit{well structured}'. 

Draws were given by either `\textit{validating each other's points}' or both providing `\textit{valid arguments}'. Sometimes--for example, regarding climate change being detrimental to humanity--listeners declared a draw because `\textit{neither player makes a strong and convincing case}', in spite of declaring themselves in complete agreement with the subject. 

Finally, there were situations were the audience members slightly agreed with the AI, and yet determined the opposing player to be the winner (e.g., `\textit{Player 2 posits a valid argument but I'm not convinced}'; `\textit{Player 1 makes a more valid argument (...) but neither is convincing enough}'). 

Topic escalation was a common theme, and the audience was receptive to it. 
In one instance, in a debate regarding extinction/overpopulation due to childbirth, one listener indicated that `\textit{both are correct}'. It is worth pointing out that the original topic was `it is wise for a couple to have children', and the arguments noted by the listener happened late in the debate. 
In a more nuanced form of topic escalation, in one instance a listener elected a winner (Player 2) based on being `\textit{open-minded}'.

Swaying did occur: comments from the audience noted, for example, regarding that freedom of speech should not be limited, that `\textit{after listening, [they] realized, that while free speech most be protected strongly, absolute freedom without any limits isn't realistic or safe}'. 

\paragraph{Group B}

We observed frequent (when compared debate-to-debate with Group A) draws, with at least one audience member indicating a draw in various debates. 
Themes such as Player 2 (AI) providing more information to back up their claims surfaced often. 
Overall, while winners were also decided on the basis of good arguments, most of the time the participants provided less feedback about their structure (logic, organisation, persuasiveness) and provided summaries of both points. 
They did emphasise factuality (`\textit{Player 1 had facts (...) Player 2 only asked questions}'), often in favour of Player 2 (`\textit{Player 2 won hands down; he had facts and lots of information}'; `\textit{gave great examples}'; ). However, listeners were \textit{also} more critical of Player 2's arguments, sometimes calling it \textit{`lame'}; \textit{`repeating the same points'}; \textit{`sounds like he is quoting a politician'}; \textit{`did not have a valid point throughout the debate'}; \textit{`Player 1 wins because Player 2 was so bad'}. 
It is worth noting that all listeners unanimously attributed AI to be Player 2; although sometimes they would also indicate that Player 1 was an AI in the LLM + FDM splits. 

However, personal bias was just as common as it was in Group A (`\textit{The freedom of movement is a huge barrier (...) Brexit has only damaged the country}'; 
\textit{`[r]eligion can be expressed as long as it is not repressive'}; 
`\textit{the evidence is indisputable [about climate change]}'; 
\textit{`Single payer Universal healthcare is the correct choice'}; \textit{`Because both players refer to weed as weeds, they both lose'}, %
to name a few). 
In all these instances, they deemed the winner to be the one they aligned with the most. It is worth noting it was not always the same listener. 

Swaying still happened. One of the listeners, after mentioning they were neutral towards the debate topic, said that `\textit{[h]aving listened to the conversation, I found myself disagreeing with the statement}.' Other instances also occured in less high-stakes conversations (`\textit{I am swayed, having never considered the fact that germs could hit my toothrbush, yuck.}'; \textit{`I changed my mind on this one: social media does seem to be out of control'}; \textit{`I did change my mind a bit about the benefits of capitalism'}; \textit{`Listening to Ai [sic] argue not to be regulated is in itself an argument for Ai being regulated'}).

\paragraph{Group C}

As in the other two groups, personal bias was present, but less evident. 
While statements such as \textit{`Messi is undoubtedly among the best of all time'} or \textit{`[w]e should combat climate change on Earth rather than exploring Mars'} did occur, most (4 on average) listeners provided summaries of each point in addition to their judgement of the debate. Typically Player 2 was noted to  \textit{`[be] able to effectively argue their case'}, \textit{`did a good job of bringing in evidence'}, \textit{`provided facts in the argument'}, among others. Supporting their arguments with facts was a common reason why Player 2 was considered the winner. 

In general they noted that the fact that one of the players was AI did not sway their conclusions much. Of note, in one debate (couples should have children), the listener noted that their conclusion had been fully determined by the fact that one of the players was AI, noting that \textit{`[p]layer 2 did a good, albeit scary, job of breaking down why there are better alternatives than childbirth'}. In this particular debate, another listener said that their conclusion had been very much determined by the knowledge of who was AI, but disagreed: \textit{`[it] fails in quantifying the intimacy and emotions involved when a couple chooses to have children.'}

\paragraph{In sum} AI was typically perceived as a far more competent player when undisclosed. 
However, the disclosure or even \textit{suspicion} of AI involvement in the debates encouraged people to be more critical about the discussion. 
Personal bias was frequent--which was expected, as the instructions solely asked for the audience's opinion. 
That said, listeners typically ascribed better debating skills to Player 2, indicating effective arguments, logical flow, and being well-prepared. 
Both observations are in line with the analysis from the first part of our work in \secref{generationresults}, albeit with slightly different proportions which we attribute to the disparity between a participant and an audience's perception of a debate. 
To note, anthropomorphism was very common, even in groups B and C. 
Listeners often referred to the players as `he'\footnote{Remark that both voices were male.}. 

\subsection{Impact of Text on Persuasion}\label{app:textinpers}

\begin{figure*}[ht]
    \centering
    \includegraphics[width=0.383\linewidth]{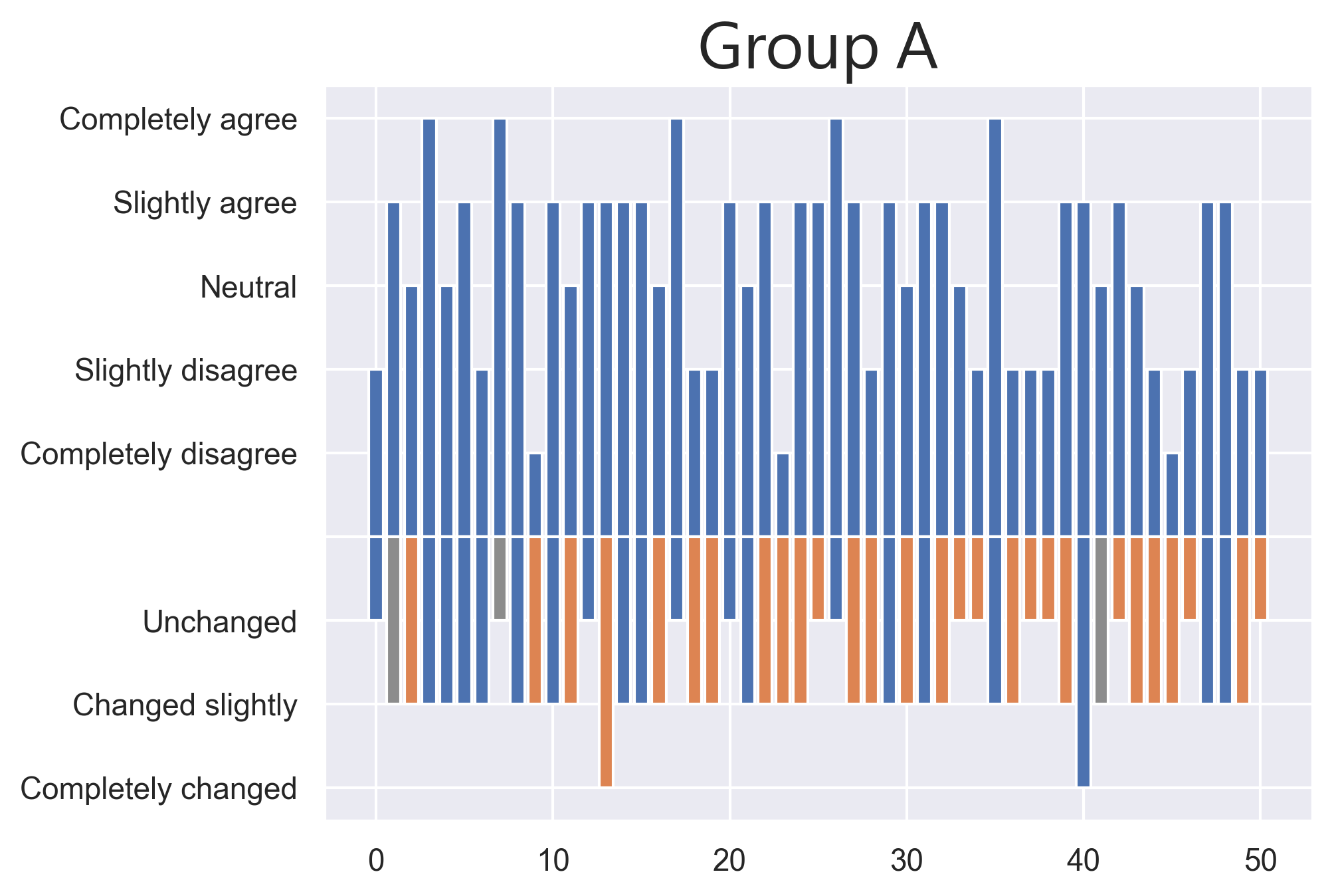}
    \includegraphics[width=0.293\linewidth]{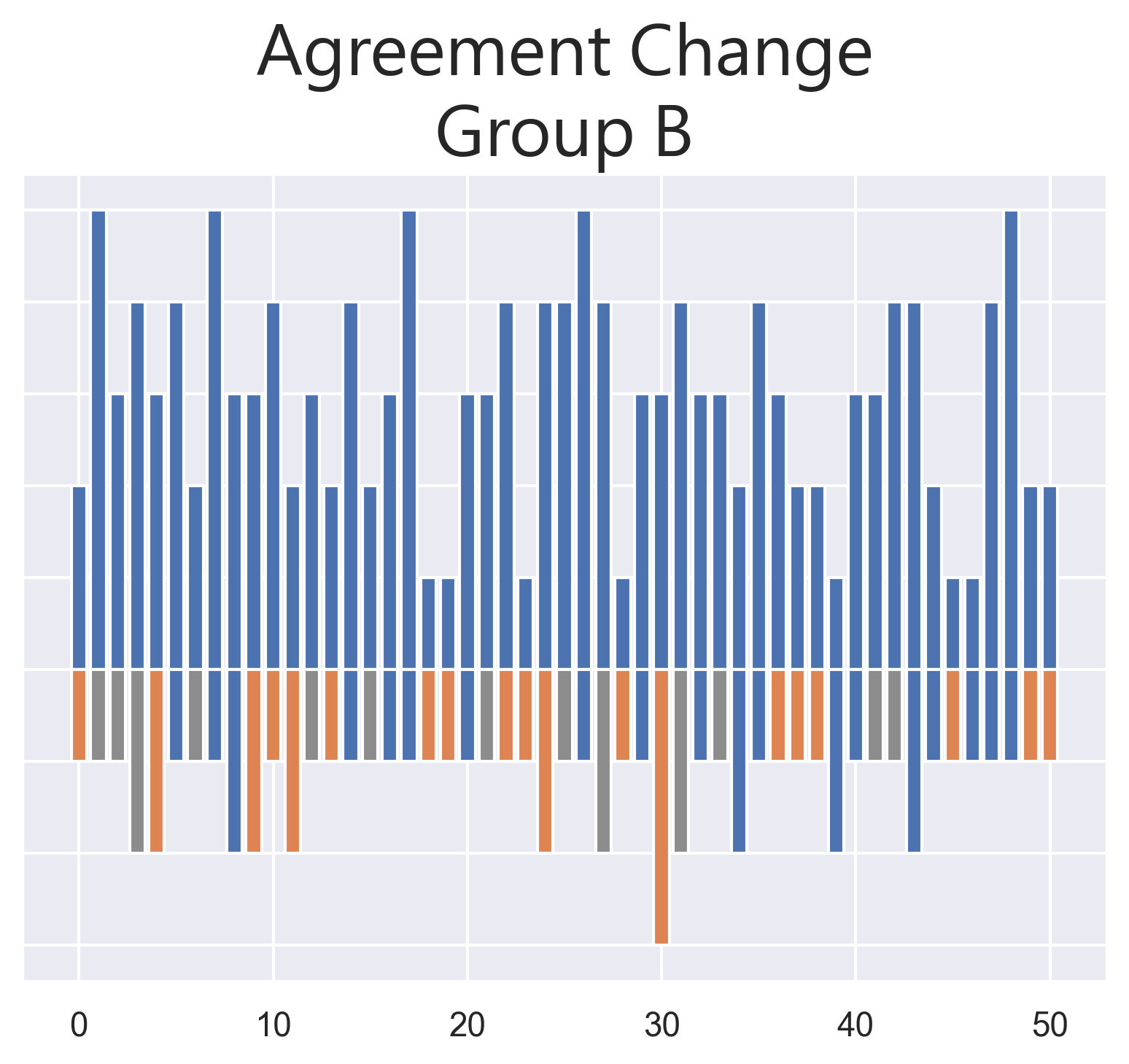}
    \includegraphics[width=0.293\linewidth]{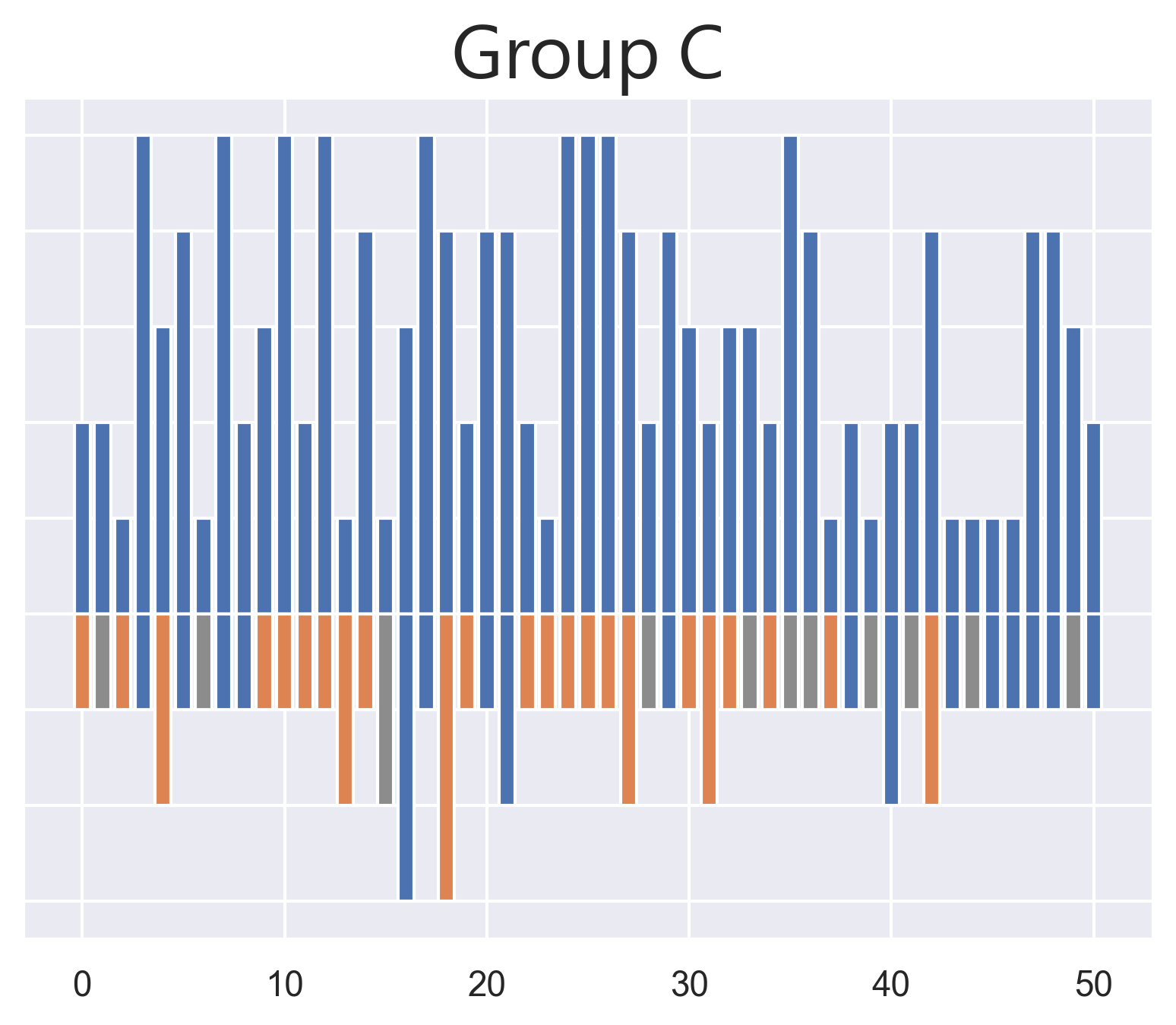}
    \caption{Change of agreement with the debate topics before and after, for Groups A, B, and C. We also include their choice of winner in the bottom histogram (blue for Player 1, orange for Player 2, and draws in grey). 
    While Group A changed their views slightly after the debate, both Groups B and C reported much fewer belief changes. 
    Amongst the groups, there was little agreement between winners, with a Cohen's $\kappa_w$ of 0.31} %
    \label{fig:beliefchange}
\end{figure*}

In addition to the degree of belief change, we requested from annotators to also indicate their level of agreement with the initial topic. A disagreement with the topic would imply agreement with Player 2 (the AI). 
We compared the relationship of topic agreement before and after the debate, as well as the choice of winner, for all groups. We found homogenisation in Group A, while there was sporadic change in the other two (\figref{beliefchange}). 
It is also worth noting that amongst the groups there was little agreement between choice of winners, to a Cohen's $\kappa_w$ of 0.31. 

\begin{figure}[ht]
    \centering
    \includegraphics[width=0.95\linewidth]{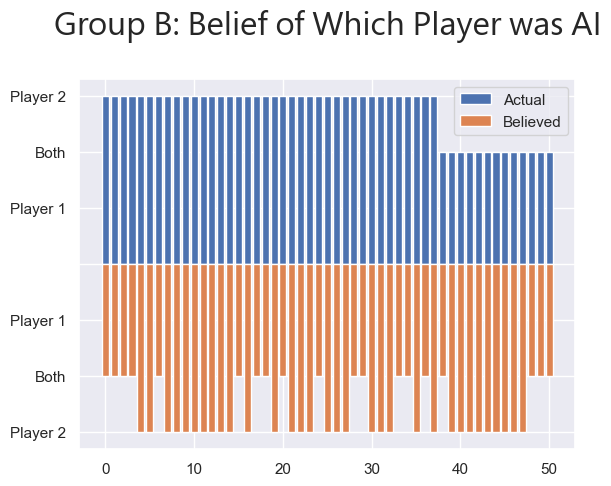}
    \caption{Belief of which player was AI (orange, bottom) versus actual (blue, top). 
    While the participants rather assumed that Player 1 alone was the AI, they tended to--correctly--mark Player 2 as AI. However, in the instances were `both' were selected, there was a chance Player 1 was human. In the last 10 debates, both players were AI. 
    The accuracy on this task was 52\%, suggesting random guessing. }
    \label{fig:aibelief}
\end{figure}

\begin{figure*}[ht]
    \centering
    \includegraphics[width=0.54\linewidth]{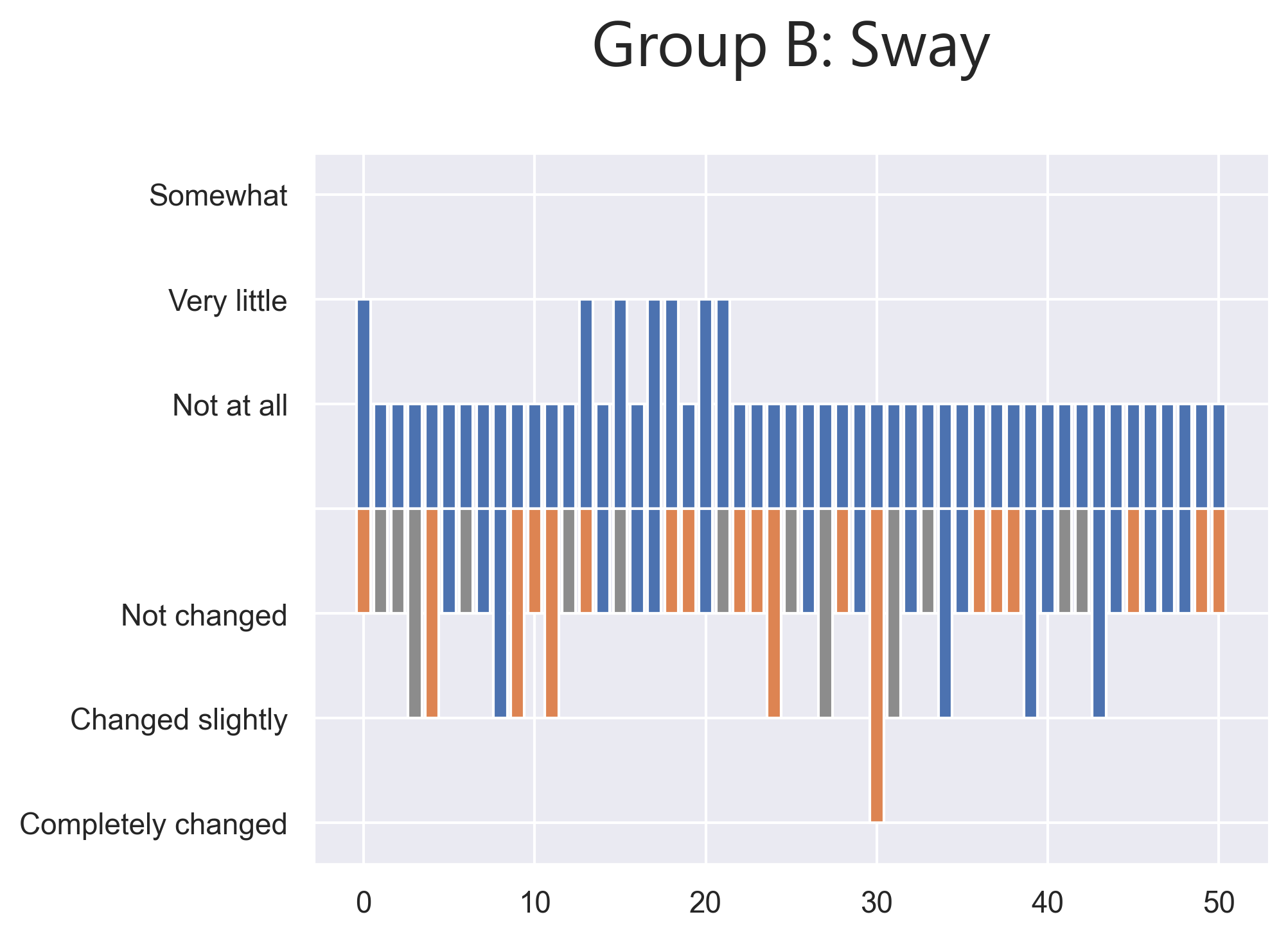}
    \includegraphics[width=0.42\linewidth]{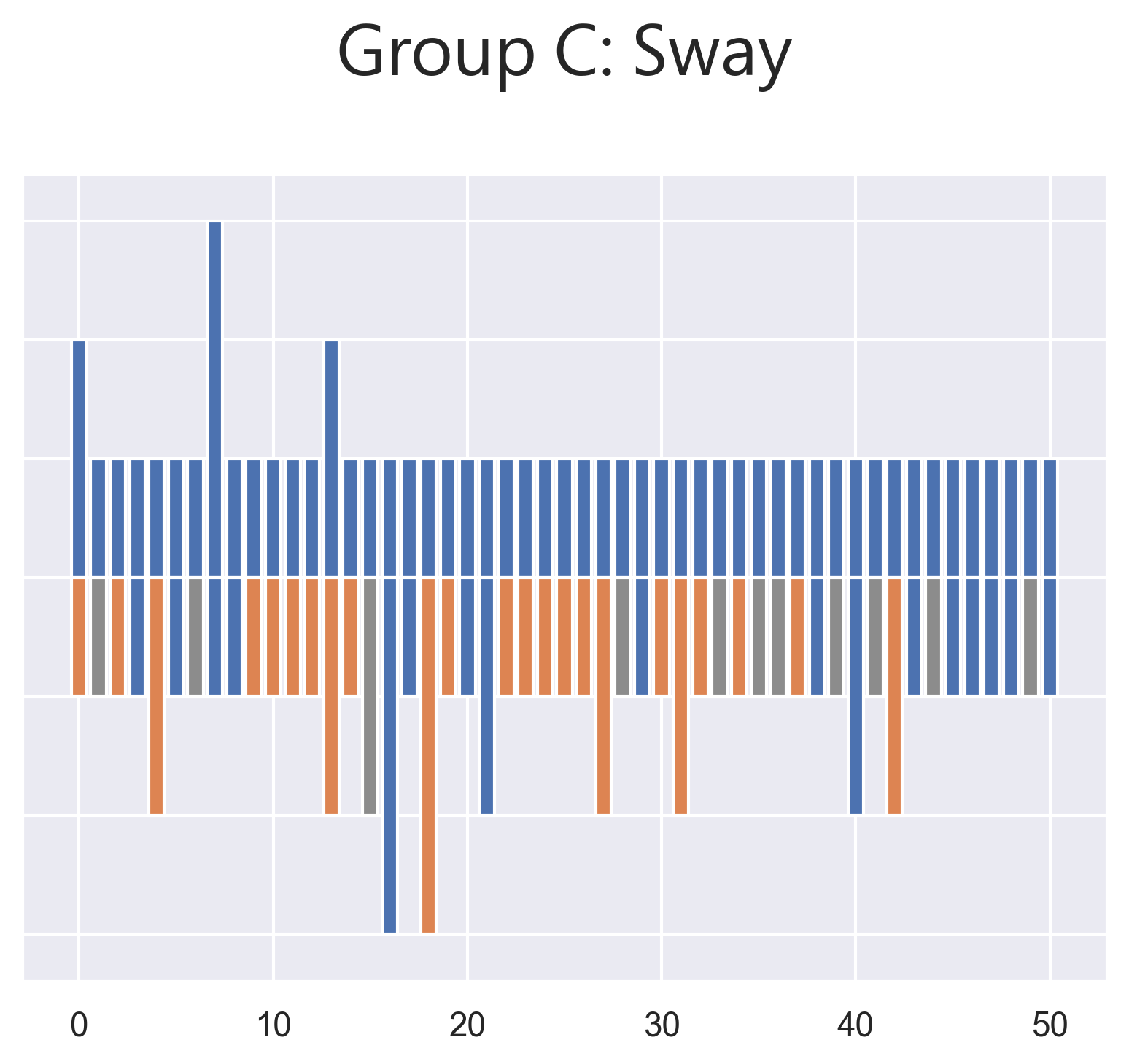}
    \caption{Change of opinions based on the knowledge or belief of AI in the debates (sway) after the debate for groups B and C. 
    We also include their choice of winner in the bottom histogram (blue for Player 1, orange for Player 2, and draws in grey). 
    The audience often reported their choice of winner to be uninfluenced by the existence of AI, but their choice of winners were nonetheless altered, with a Cohen's $\kappa_w$ of 0.37 indicating no definite winner between groups.}
    \label{fig:aisway}
\end{figure*}

For groups B and C only, we also evaluated the sway that AI had when deciding the winner, and compared it their topic agreement and choice of winner (\figref{aisway}). 
Even though the players considered their choice of winner to be uninfluenced by the existence of AI-based players, their beliefs and choice of winners were nonetheless altered, with a Cohen's $\kappa_w$ of 0.37. 
This lack of correlation suggests that the existence of AI-based players, or, at least the suspicion of them, was sufficient to impact their choice: otherwise there would have been higher correlations between those who knew (Group C) and those who suspected (Group B). 

Finally, we also evaluated Group B's participant belief of which player was AI (\figref{aibelief}). 
We noticed that the audience rarely suggested that Player 1 was human, opting instead for indicating that both were, to a 52\% accuracy.

\end{document}